\documentclass[lettersize,journal]{IEEEtran}
\usepackage{amsmath,amsfonts}
\usepackage{algorithm}
\usepackage{array}
\usepackage{placeins}
\usepackage{textcomp}
\usepackage{ltablex}
\usepackage{threeparttable}
\usepackage{stfloats}
\usepackage{url}
\usepackage{verbatim}
\usepackage{graphicx}
\usepackage{cite}
\usepackage{subfigure}
\usepackage{setspace}
\usepackage{geometry} 
\usepackage{comment}
\usepackage{xcolor}
\usepackage{rotating}
\usepackage[justification=centering]{caption}
\usepackage{multirow} 
\usepackage{hyperref}
\usepackage{makecell}
\newcommand{\etal}{\textit{et al.}}
\newcommand{\eg}{\textit{eg}}
\usepackage[most]{tcolorbox}
\graphicspath{{./figures/}} 
\usepackage{boldline} 
\usepackage{ulem}
\usepackage{color}
\usepackage[activate]{microtype}
\usepackage{svg}

\usepackage{authblk}
\usepackage{booktabs}

\usepackage{colortbl}
\usepackage[separate-uncertainty = true,multi-part-units = repeat]{siunitx}
\usepackage{array,amsfonts,amsmath}
\usepackage{caption}
\usepackage[font=small,labelfont=bf,
   justification=justified,
   format=plain]{caption} 
\usepackage{tablefootnote}
\usepackage[english]{babel}
\usepackage[utf8]{inputenc}
\usepackage{adjustbox}
\usepackage{pdflscape}
\usepackage{afterpage}
\usepackage{sidecap}
\usepackage{algpseudocode}
\usepackage{lscape}
\usepackage{epstopdf}
\usepackage{lipsum}

\usepackage{footnotehyper}
\definecolor{mygray}{gray}{.9}
\usepackage{bm}
\usepackage{microtype}
\begin{document}

\title{Transformers in Remote Sensing: A Survey}

\author{ Abdulaziz Amer Aleissaee\textsuperscript{*}, Amandeep Kumar\textsuperscript{*},  Rao Muhammad Anwer, Salman Khan, Hisham Cholakkal, Gui-Song Xia  and Fahad Shahbaz khan
\IEEEcompsocitemizethanks{\IEEEcompsocthanksitem A.A. Aleissaee, A. Kumar, R. Anwer, S. Khan, H. Cholakkal and F. Khan are with MBZ University of Artificial Intelligence, UAE.\protect\\
Gui-Song Xia is with Wuhan University, China.\protect\\
E-mail: \{firstname.lastname\}@mbzuai.ac.ae}}

\maketitle
\begin{abstract}
Deep learning-based algorithms have seen a massive popularity in different areas of remote sensing image analysis over the past decade. Recently, transformers-based architectures, originally introduced in natural language processing, have pervaded computer vision field where the self-attention mechanism has been utilized as a replacement to the popular convolution operator for capturing long-range dependencies. Inspired by recent advances in computer vision, remote sensing community has also witnessed an increased exploration of vision transformers for a diverse set of tasks. Although a number of surveys have focused on transformers in computer vision in general, to the best of our knowledge we are the first to present a systematic review of recent advances based on transformers in remote sensing. Our survey covers more than 60 recent transformers-based methods for different remote sensing problems in sub-areas of remote sensing: very high-resolution (VHR), hyperspectral (HSI) and synthetic aperture radar (SAR) imagery. We conclude the survey by discussing different challenges and open issues of transformers in remote sensing. Additionally, we intend to frequently update and maintain the latest transformers in remote sensing papers with their respective code at: \url{https://github.com/VIROBO-15/Transformer-in-Remote-Sensing}
$\let\thefootnote\relax\footnote{* Authors contributed equally}$
\end{abstract}
\section{Introduction}
Remote sensing imaging technology has significantly advanced in the last decades. Modern airborne sensors provide a large coverage of the Earth surface  with improved spatial, spectral and temporal resolutions, thereby playing a crucial role in numerous research areas, including ecology, environmental science, soil science, water contamination, glaciology, land surveying and analysis of the crust of the Earth. Automatic analysis of remote sensing imaging brings unique challenges such as, data are generally multi-modal (\eg , optical or synthetic aperture radar sensors), located in the geographical
space (geo-located) and typically on a global-scale with ever growing data volumes. 

Deep learning, especially convolutional neural networks (CNNs) has dominated many areas of computer vision, including object recognition, detection and segmentation. These networks typically take an RGB  image as an input and perform a series of convolution, local normalization and pooling operations. CNNs typically rely on a large amount of training data and the resulting pre-trained models are then utilized as generic feature extractors for a variety of downstream applications. The success of deep learning-based techniques in computer vision has also inspired the remote sensing community with significant advances being made in many remote sensing tasks, including hyperspectral image classification, change detection and very high-resolution satellite instance segmentation.  

One of the main building blocks in CNNs is the convolution operation which captures local interactions between elements (\eg, contour and edge information) in the input image. CNNs encode biases such as, spatial connectivity and translation equivariance. These charactertistics aid in constructing generalizable and efficient architectures. However, the local receptive field in CNNs limits modeling long-range dependencies in an image (\eg, distant part relationships). Moreover, convolutions are content-independent as the convolutional filter weights are stationary with same weights applied to all inputs  regardless of their nature. Recently, vision transformers (ViTs) \cite{dosovitskiy2021an} have demonstrated impressive performance across a variety of tasks in computer vision. ViTs are based on the self-attention mechanism  that effectively captures global interactions by learning the relationships between the elements of a sequence. Recent works \cite{MuzammalNeurIPS21,ParkICLR22} have shown that ViTs possess  content-dependent
long-range interaction modeling capabilities and can flexibly adjust their receptive fields to counter nuisances in data and learn effective feature representations. As a result, ViTs and their variants have been successfully utilized for many computer vision tasks, including classification, detection and segmentation. 

\begin{figure*}[t!]
    \centering
		\includegraphics[width=1.0\textwidth]{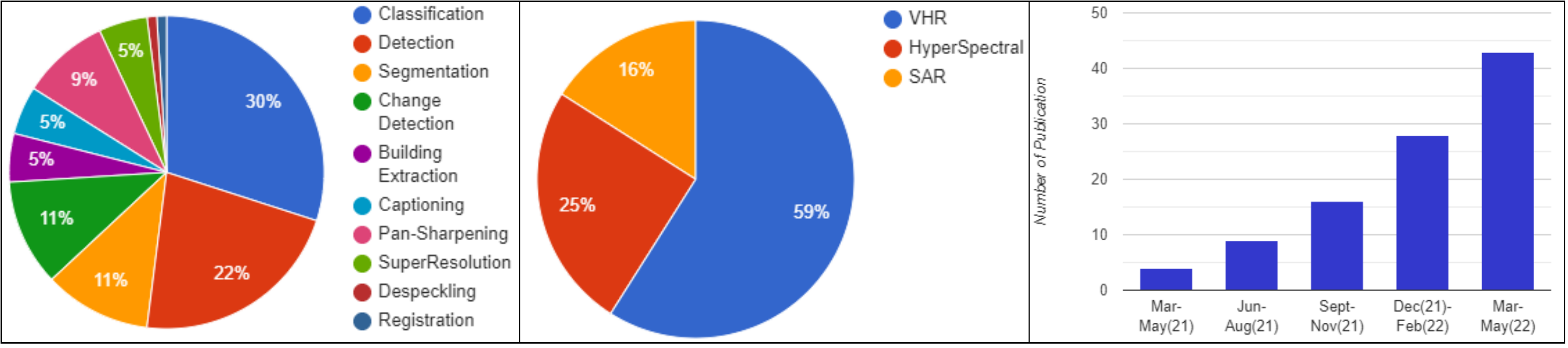}
	\caption{Recent transformers-based techniques in the remote sensing imaging. On the left and middle: Pie-charts are representing statistics of the articles covered in this survey in terms of different remote sensing imaging problems and data type representations. On the right: we show a plot illustrating the consistent increase in number of papers recently. }
	\label{fig:ViT-graph}
\end{figure*}

Following the success of ViTs in computer vision, remote sensing community has also witnessed a significant growth (see Fig.\ref{fig:ViT-graph}) in the employment of transformers-based frameworks in many tasks such as, very high-resolution image classification, change detection, pan sharpening, building detection and image captioning. This has started a new wave of promising research in remote sensing with different approaches utilizing either ImageNet pre-training \cite{bazi2021vision, hao2022two, ma2022homo} or performing remote sensing pre-training \cite{Wang2022empirical} with vision transformers. Similarly, there exist approaches in literature that are based on pure transformers design \cite{hong2021spectralformer, liu2022dss} or utilize a hybrid approach \cite{9762303, 9766028, jia2022multiscale} based on both transformers and CNNs. It is therefore becoming increasingly challenging to keep pace with the recent progress due to the rapid influx of transformers-based methods for different remote sensing problems. In this work, we review these advances and present an account of recent transformers-based approaches in the popular field of remote sensing. To summarize, our main contributions are the following:

\begin{itemize}
\item We present a holistic overview of applications of transformers-based models in remote sensing imaging. To the best of our knowledge, we are the first to present a survey on transformers in remote sensing, thereby bridging the gap between recent advances in computer vision and remote sensing in this rapidly growing and popular area. 

\item We present an overview of both CNNs and transformers, discussing their respective strengths and weaknesses. 
\item We present a review of more than 60 transformers-based research works in the literature to discuss the recent progress in the field of remote sensing.
\item Based on the presented review, we discuss different challenges and research directions on transformers in remote sensing.
 
\end{itemize}

The rest of the paper is organized as follows: Section~\ref{Related_Work} discusses other related surveys on remote sensing imaging. In Section~\ref{imaging-types}, we present an overview of different imaging modalities in remote sensing, whereas Section~\ref{sec:ViT} provides a brief overview of CNNs and vision transformers. Afterwards, we review advances with respect to transformers-based approaches in very high-resolution (VHR) imaging (Section~\ref{sec:VHR}), hyperspectral image analysis (Section~\ref{sec:HIA}) and synthetic aperture radar (SAR) in Section~\ref{sec:SARS}. In Section~\ref{VHR_discussion}, we conclude our survey and discuss potential future research directions.

\section{Related Work}\label{Related_Work}

In the literature, several works have performed a review of machine learning techniques for remote sensing imaging in the past decade.
Tuia \textit{et al.} \cite{Tuia11} compare and evaluate different active learning algorithms for supervised remote sensing image classification task. The work of \cite{Gustavo13} focuses on the problem of hyperspectral image classification and reviews recent advances in relation to machine learning and vision techniques. Zhu \textit{et al.} \cite{zhu2017deep} present a comprehensive review of utilizing deep learning techniques for remote sensing image analysis. Their work provides a comprehensive review of the existing approaches along with describing a  list of resources about deep learning in remote sensing. Ma \textit{et al.} \cite{ma2019deep} review major deep learning concepts in remote sensing with respect to image resolution and study area. To this end, their work studies different remote sensing tasks such as, image registration, fusion, scene classification and object segmentation. 

Recently, transformers-based approaches have witnessed a significant surge within the computer vision community, following the breakthrough from transformers-based models \cite{vaswani2017attention} in natural language processing (NLP). Khan \textit{et al.} \cite{KhanVitSurvey} present an overview of the transformers models in vision with emphasis on recognition, generative modeling, multi-modal, video processing and low-level vision tasks. Shamshad \textit{et al.} \cite{Shamshad2022} survey the use of transformers models in medical imaging, focusing on different medical imaging tasks such as, segmentation, detection, reconstruction, registration and clinical medical report generation. The work of \cite{Selva2022} presents an overview of the growing trend of using transformers to model video data. Their work also compares the performance of vision transformers on different video tasks such as, action recognition. 

Different from the aforementioned surveys, our work presents a review of recent advances of transformers-based approaches in the popular area of remote sensing. To the best of our knowledge, this is the first survey presenting a comprehensive account of transformers in remote sensing, particularly dedicated to progress in very high-resolution, hyperspectral and synthetic aperture radar image analysis. 

\begin{figure}[!t]
    \centering
		\includegraphics[width=\linewidth]{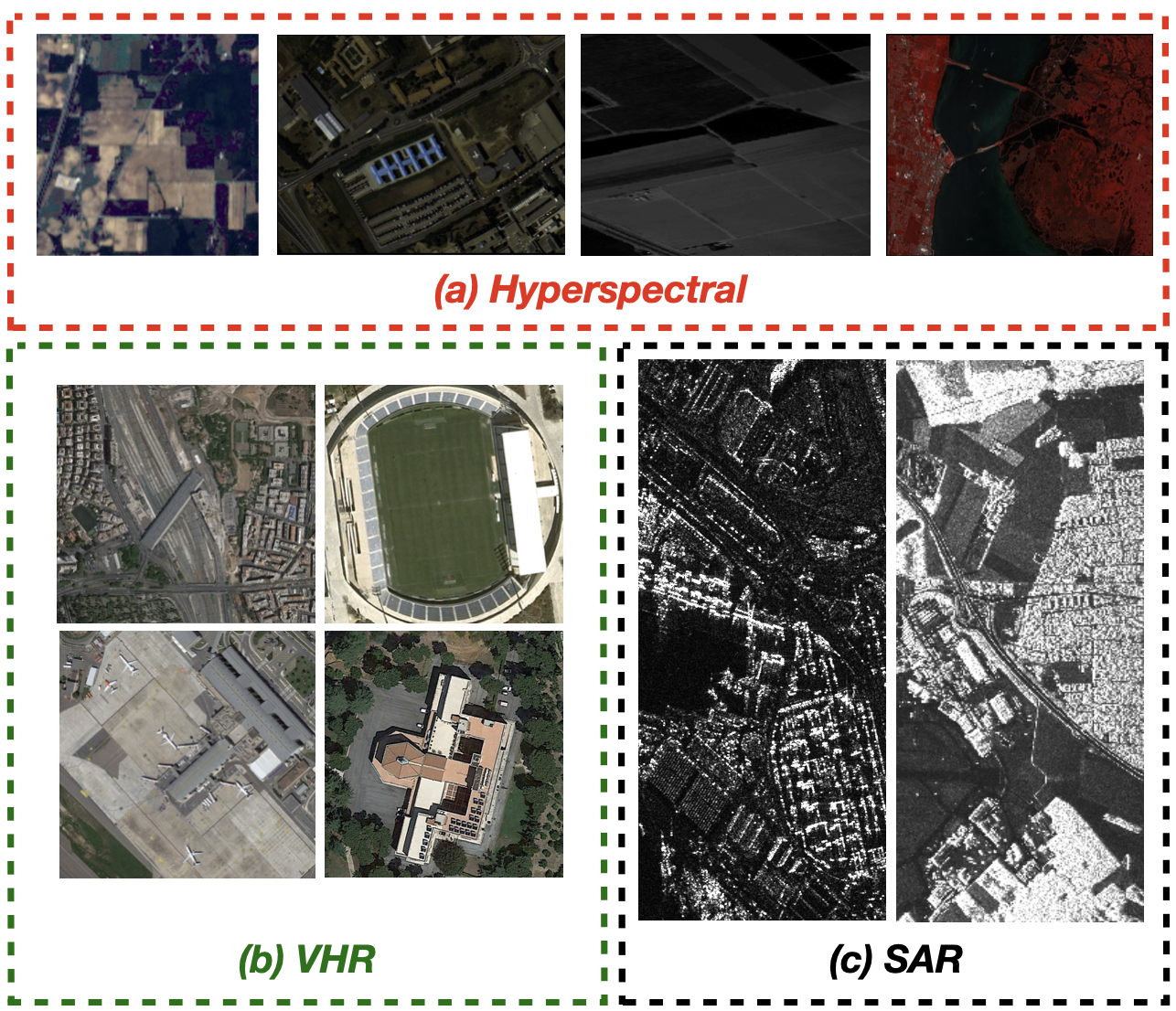}
	\caption{Example hyperspectral (a), VHR (b) and SAR (c) images from different datasets.}
	\label{fig:Images}
\end{figure}

{\section{Remote Sensing Imaging Data}\label{imaging-types}} 
Remote sensing imagery is generally acquired from a range of sources as well as data collection techniques. Remote sensing image data can be typically characterised by their  spatial, spectral, radiometric, and temporal resolutions.  Spatial resolution refers to each pixel size within an image  along with the area of the surface of the Earth represented by that corresponding pixel. Spatial resolution characterizes the small and fine-detailed features in an imaging scene that can be separated. Spectral resolution refers to the capability of the sensor to collect information about the scene by discerning finer wavelengths, with having more narrower bands (e.g., 10 nm). On the other hand, radiometric resolution characterizes the extent of the information in each pixel, where a larger dynamic range for a sensor implies more details are to be discerned in the image. The temporal resolution refers to the time it takes between consecutive images of the same location on ground acquired by the sensor. Here, we briefly discuss commonly utilized remote sensing imaging types with few examples shown in Fig.\ref{fig:Images}. \\

\textbf{Very High-resolution Imagery:} In recent years, the emergence of very high-resolution (VHR) satellite sensors has paved the way towards yielding the higher spatial resolution imagery  beneficial for land use change detection, object-based image analysis (object detection and instance segmentation), precision agriculture farming (e.g., management of crops, soil and pests) and emergency responses. Furthermore, these recent advances in sensor technology along with new deep learning-based techniques allow the usage of VHR remote sensing imagery to analyze the biophysical as well as biogeochemical processes both in coastal and inland waters. Nowadays, optical sensors produce panchromatic and multispectral imagery of the Earth’s surface at a much finer spatial resolutions (e.g., 10 to 100 cm/pixel). 

\textbf{Hyperspectral Imagery:}
Here, each pixel in the scene  is captured using  continuous spectrum of light with fine wavelength resolutions. The continuous spectrum  extends  wavelengths beyond the visible spectrum and include wavelengths from ultraviolet (UV) to infrared (IR).  
Generally, spectral resolution of  hyperspectral images are expressed using the wave number along with the nanometers (nm). The most popular continuous spectrum used for measuring the pixels are mid-infrared, that is near infrared and visible wavelength bands. 
In order to acquire hyperspectral imagery, there are different electromagnetic measurements such as, Raman spectroscopy,  X-ray spectroscopy, Terahertz spectroscopy, 3D ultrasonic imaging,  magnetic resonance and confocal laser microscopy scanners that can measure the entire emission spectrum for each pixel at a specific excitation wavelength. 
The hyperspectral images have high dimensionality and strong resolving power for fine spectra. The imagery offers a wide range of applications, including in environmental science \cite{teng2013investigation} and mining \cite{notesco2014mineral}. Different from regular images that contain only the primary colors (red, green, blue) within the visible spectrum, hyperspectral images are rich in spectral information that can reflect the physical structure and chemical composition of the item of interest. In remote sensing, automatically analyzing hyperspectral imagery is an active  research topic.

\textbf{Synthetic Aperture Radar Imagery:}
A large amount of synthetic aperture radar (SAR) images are produced by Earth observation satellites every day through emission and reception of electromagnetic signals. In the past decades, SAR images have gained popularity  due to their higher spatial resolution, all-weather capability, de-speckling tools such as, CAESAR along with recent advances in the SAR specific image processing. SAR imagery can be used for numerous applications, including geographical localization, object detection, functionalities of basic radars, and geophysical features estimation of complex setting such as, roughness, moisture content, and density. Further, SAR imagery can be used for disaster management (oil slick detection, ice tracking), forestry and hydrology. \\

\section{From CNNs to Vision Transformers}\label{sec:ViT}
In this section, we first present a brief overview of CNNs and then provide a brief description of vision transformers recently utilized for different vision tasks.

\subsection{Convolutional Neural Networks}\label{CNN}
Convolutional neural networks (CNNs) have dominated a variety of computer vision tasks, including image classification \cite{krizhevsky2012imagenet} and object detection \cite{ren2015faster}. CNNs are typically made up of series of two main parts: convolutional and pooling layers. The convolutional layer produces feature maps by convolving the local region in the input with a set of kernels. These features are subjected to a non-linear function with the same process repeated for each convolutional layer. In CNNs, the pooling layer carries out a downsampling operation (typically utilizing the max or mean operation) to feature maps. In different existing CNN architectures, the convolutional and pooling layers are followed by a set of fully connected layers, where the last fully connected layer is the softmax computing each object category score. \\
\textbf{Popular CNN Backbones:} Here, we briefly discuss different popular CNN backbone architectures in literature. \\
\textit{AlexNet:} Krizhevsky \textit{et al.} \cite{krizhevsky2012imagenet} propose a CNN architecture, named AlexNet, for image classification task. AlexNet comprises five convolutional layers followed by three fully-connected layers. The proposed network architecture utilizes Rectified Linear Units (ReLU) for training efficiency. The network contains 60 million parameters and 500,000 neurons with network training performed on the large-scale ImageNet dataset \cite{5206848}. Different data augmentation techniques are employed to increase the training set. In the ImageNet 2012 competition, AlexNet achieved competitive performance with top-1 and top-5 error rates of 39.7\% and 18.9\%, respectively. \\
\textit{VGGNet:} Different from AlexNet, Simonyan and Zisserman \cite{simonyan2014very} introduce an architecture, named VGGNet that comprises 16 layers in total. The network takes an input image of 224 $\times$ 224 size and has around 138 million parameters. It uses different data augmentation techniques, including scale jittering during network training. The VGGNet architecture comprises convolution layers of 3 $\times$ 3 filter, where the receptive fields are convolved at each pixel with a stride of one pixel. The VGGNet contains multiple pooling layers, performing spatial pooling over 2 $\times$ 2 windows with a stride of two pixels. Further, VGGNet contains two fully connected layers followed by a softmax for yielding output predictions. The VGG architecture achieved top classification accuracy on the 2014 ImageNet classification challenge. \\  
\textit{ResNet:} Different from AlexNet and VGGNet, He \textit{et al.} \cite{he2016deep} introduce residual neural networks (ResNet) that stacks residual blocks to build a network. ResNet provides a residual learning approach for training networks that are much deeper than their previously utilised counterparts. Instead of learning un-referenced functions, it explicitly reformulates the layers as learning residual functions with reference to the layer inputs. Extensive empirical evidence demonstrates that residual networks are easier to optimize with improved accuracy from higher depth. 

The development of CNN-based architectures has led to rise of novel techniques, improved hardware (e.g. GPUs and TPUs), better optimization methods and many open-source libraries. Interested readers can go through the survey papers related to CNN methods for remote sensing \cite{zhu2017deep,ma2019deep}. Previous works have analyzed that CNNs are able to capture image-specific inductive bias which increases their effectiveness in learning better feature representations. However, CNNs do not capture long-range dependencies that aids to enhance expressivity of the representations. Next, we briefly present vision transformers that are capable of modelling long-range dependencies in the images.

\begin{figure*}[t!]
    \centering
		\includegraphics[width=1.0\textwidth]{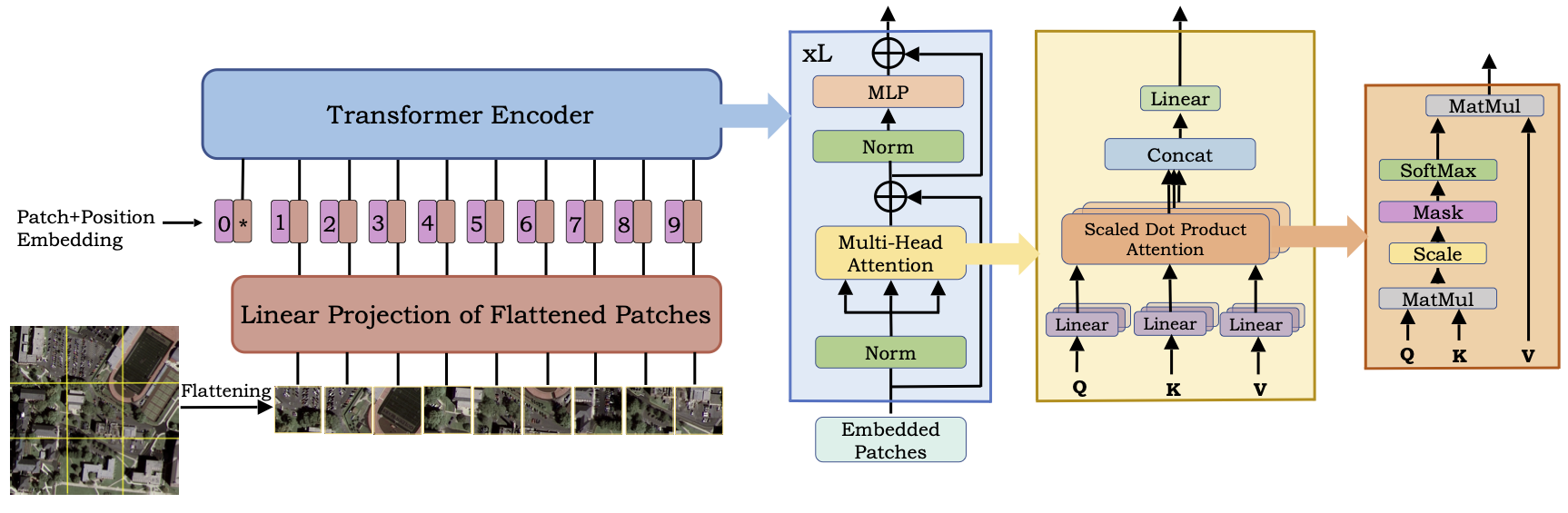}
	\caption{ The Vision Transformer’s architecture is shown on the left and the encoder block’s specifications are shown on the right. The
input image is first divided into patches. These are then projected (after flattening) into a feature space, where a
transformer encoder analyses them to create the classification output. Adapted from \cite{dosovitskiy2021an} and \cite{Shamshad2022}.}
	\label{fig:ViT-fig}
\end{figure*}

\begin{figure*}[t!]
    \centering
		\includegraphics[width=1.0\textwidth]{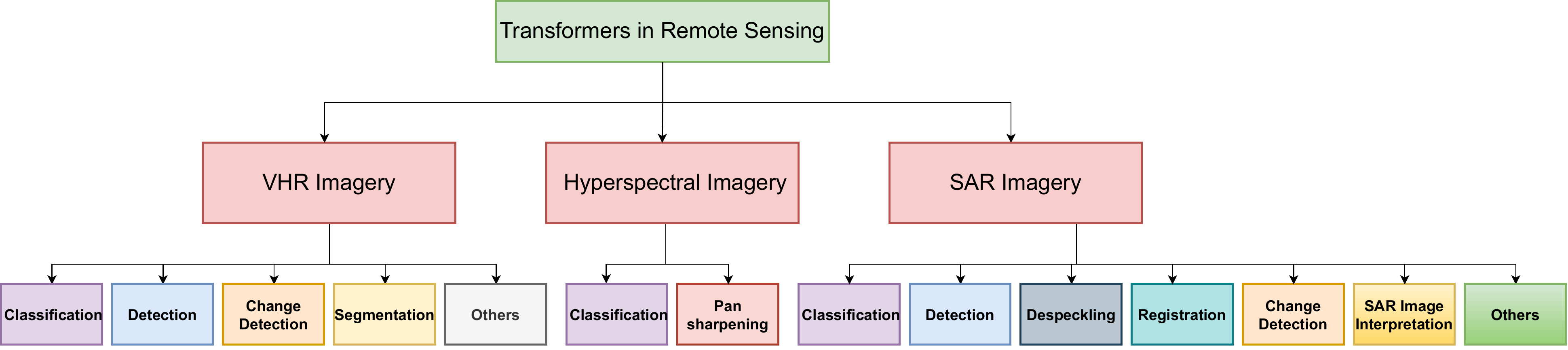}
	\caption{The taxonomy of transformers in VHR, hyperspectral and SAR imagery with a variety of tasks such as, classification, detection, segmentation, pan sharpening and change detection. }
	\label{fig:ViT-blockDG}
\end{figure*}

\subsection{Vision Transformers}\label{ViT}
Recently, transformers-based models have achieved promising results across many computer vision and natural language processing (NLP) tasks. Vaswani \etal \cite{vaswani2017attention} first introduce transformers as an  attention-driven model for  machine translation application. To capture the long-range dependencies, transformers use self-attention layers instead of traditional recurrent neural network that struggles to encode such dependencies between the elements of a sequence. 

To effectively capture the long-range dependencies within an input image, the work of \cite{dosovitskiy2021an} introduces vision transformers (ViTs) for image recognition task, as shown in Fig.\ref{fig:ViT-fig}. ViTs \cite{dosovitskiy2021an} interpret an image as a  sequence of patches and process it via a conventional transformers encoder similar to those used in NLP tasks. The success  of ViTs in generic visual data have sparked the interest not only in different areas of computer vision but also in the remote sensing community, where  a number of  ViT-based techniques have been explored in recent years for various  tasks. 

Next, we briefly describe the key component of self-attention within transformers.\\
\noindent\textbf{Self-Attention:}
The self-attention mechanism has been an integral component of transformers as it captures the long-range dependencies and encodes the interaction between all of the sequences tokens (patch
embedding). The key idea of self-attention is to learn self-alignment, that is to update the token by aggregating global knowledge from all the other tokens in the sequence\cite{bahdanau2014neural}.  Given a 2D image $x \in \mathbb{R}^{H \times W \times C }$, the process starts with flattening the image into a series of 2D patches $x_pat \in \mathbb{R}^{M \times (P^2C)}$, where $C$ represents number of channels, $H$ and $W$ represents height and width of the image, $P \times P$ is the dimension of each individual patch, and $M = HW/P^2$ represents the total number of patches. A learnable linear projection layer of $E$ dimension is used to project these flattened patches and can be showed as a matrix $X \in \mathbb{R}^{N \times E}$. The aim of the self-attention is to  apprehend interaction among all the $M$ embeddings, that is achieved by introducing the three learnable weight matrices to modify input $X$ into queries (as $W^Q \in \mathbb{R}^{E\times E_q}$), keys (as $W^K \in \mathbb{R}^{E\times E_k}$) and values (as $W^V \in \mathbb{R}^{E\times E_v}$), where $E_q = E_k$. The sequence $X$ is first  projected onto these weight matrices to obtain $K = XW^K$, $V = XW^V$ and $Q = XW^Q$. The relative attention matrix $A \in \mathbb{R}^{M \times M}$ is 
\begin{equation}
\bm{Z} = softmax(\frac{QK^T}{\sqrt{E_q}})V
\end{equation}

\noindent\textbf{Masked Self-Attention:}
All entities are attended to the usual self-attention layer. These self-attention blocks used in the decoder for the transformers model \cite{vaswani2017attention}, which is trained to anticipate the next entity in the sequence, are masked to prevent attending to the subsequent future entities. This task is performed by an element-wise multiplication operation with a mask $\mathbf{M} \in \mathbb{R}^{n\times n}$, where $\mathbf{M}$
is an upper-triangular matrix. Here, masked self-attention is represent by 
\begin{equation}
softmax(\frac{QK^T}{\sqrt{d_q}} \circ \mathbb{M})
\end{equation}

where $\circ$ represents the Hadamard product. In masked self-attention, the attention ratings of future entities are set to zero when predicting an entity in the sequence. \\
\noindent\textbf{Multi-Head Attention:}
Multi-head attention (MHA) comprises multiple self-attention blocks concatenated simultaneously channel-wise, in order to capture different complex interactions between different sequence of embeddings. Each of the head of the multi-head self-attention has its own learnable weight matrices represented as $W^{Q_i}$, $W^{K_i}$ and $W^{V_i}$, where $i = 0 \cdot \cdot \cdot \cdot \cdot \cdot(h -1)$ were $h$ denotes the number head in multi-head self-attention. Hence, we can express,
\begin{equation}
MHA(Q,K,V) = [Z_0,...,Z_{h-1}]W^O
\end{equation}
where output of each head is concatenated to form single matrix $B \in \mathbb{R}^{M \times h\cdot E_v}$, whereas $W^O \times \mathbb{R}^{h.E_v\times M}$ computes the linear transformation of the heads.\\
\textbf{Popular Transformers Backbones:} Here, we briefly discuss some recent transformers-based backbones. \\
\textit{ViT:} The work of \cite{dosovitskiy2021an} introduces an
architecture, where a pure transformer is utilized
directly to a sequence of image patches for the task of image classification. The ViT architecture design does not employ image-specific inductive biases (e.g.,  translation equivariance and locality) and the pre-training is performed on large-scale ImageNet-21k or JFT-300M dataset.\\
\textit{Swin:} Liu \textit{et al.} \cite{liu2021swin} improve over ViT design by introducing an architecture that produces hierarchical feature representation. Swin transformer has linear computational complexity 
with respect to input image size, where the efficiency is achieved by restricting the self-attention computation to non-overlapping local windows while enabling cross-window connection. \\
\textit{PVT:} The work of \cite{Wang2021PVT} introduces a pyramid vision transformer (PVT) architecture, to perform pixel-level dense prediction tasks. The PVT architecture utilizes a progressively shrinking pyramid and a spatial-reduction attention layer for producing high-resolution multi-scale feature maps. The PVT backbone has shown to achieve impressive performance on object detection and segmentation tasks, compared to its CNNs counterpart with similar number of parameters.

Transformers offer unique characteristics that are useful for different vision tasks. Compared to the convolution operation in CNNs where static filters are computed, filters in self-attention are dynamically calculated. Furthermore, permutations and changes in the number of input points have little effect on self-attention. Recent studies \cite{MuzammalNeurIPS21,ParkICLR22} have explored different interesting properties of vision transformers and compare them with CNNs. For instance, the recent work of \cite{MuzammalNeurIPS21}  shows that vision transformers are more robust to severe occlusions, domain shifts and pertubations. Next, we present a review of transformers in remote sensing based on the taxonomy shown in Fig.\ref{fig:ViT-blockDG}.

\section{Transformers in VHR Imagery}\label{sec:VHR}
Here, we review transformers-based approaches utilized to address different problems in very-high resolution (VHR) imagery.

\subsection{Scene Classification}\label{MHSA}
\begin{figure}[!t]
    \centering
		\includegraphics[width=\linewidth]{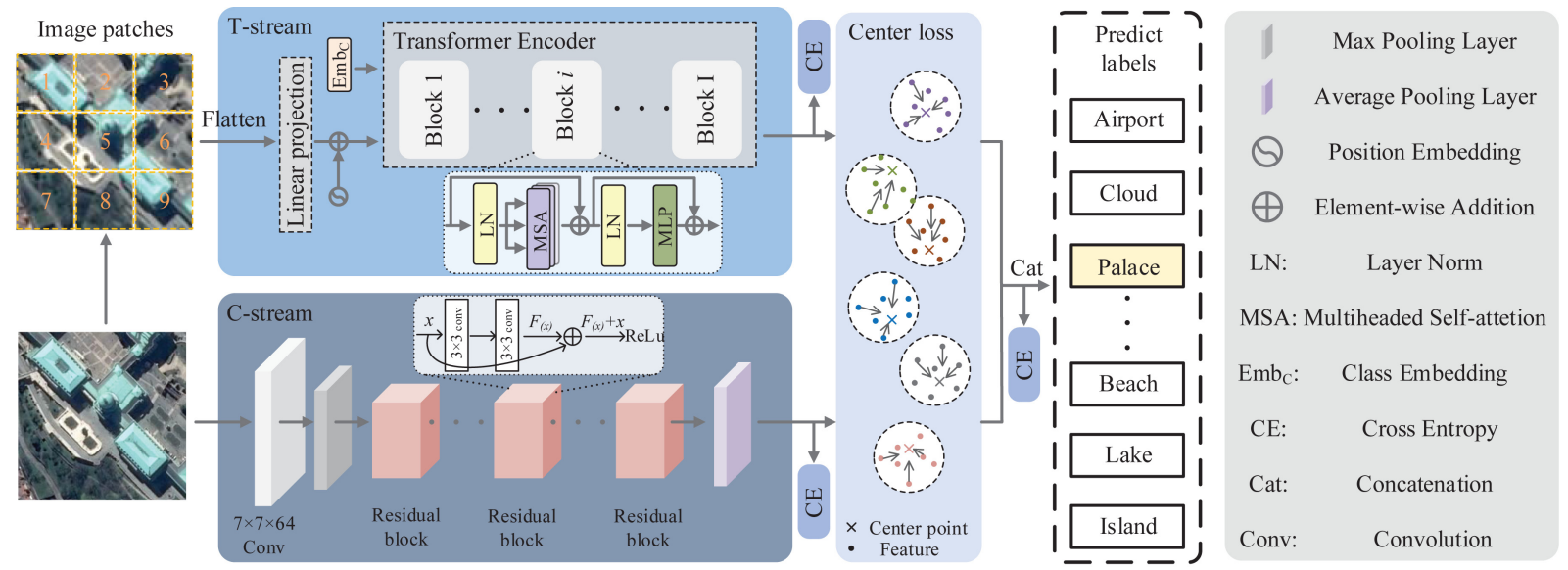}
	\caption{The CTNet architecture comprising two modules: the ViT stream (T-stream) and the
CNNs stream (C-stream). The T-stream and C-stream are designed to  capture semantic features and the local structural information. Figure is from \cite{deng2021cnns}. Best viewed zoomed in. }
	\label{fig:CT_Net_VHR}
\end{figure}

Remote sensing scene classification is a challenging problem, where the task is to automatically associate a semantic category label to a given high-resolution image comprising ground objects and different land cover types. Among the existing vision transformers-based VHR scene classification approaches, Bazi \etal \cite{bazi2021vision} explore the impact of standard vision transformers architecture of \cite{dosovitskiy2021an} (ViT) and investigate different data augmentation strategies for generating addition data. In addition, their work also evaluate the impact of compressing the network by pruning the layers while maintaining the classification accuracy. The work of  \cite{deng2021cnns} introduces a joint CNN-transformers framework, where there is one stream of CNNs and another stream of ViT, as shown in Fig.\ref{fig:CT_Net_VHR}. The features from the two streams are concatenated and the entire framework is trained using a joint loss function, comprising cross-entropy and center losses, to optimize the two-stream architecture. Zhang \etal \cite{zhang2021trs} introduce a framework, called Remote Sensing Transformer (TRS), that strives to combine the merits of CNNs and transformers by replacing the spatial convolutions with multi-head self-attention. The resulting multi-head self attention bottleneck has fewer parameters and is shown to be effective compared to other bottlenecks. The work of \cite{hao2022two} introduces a two-stream Swin transformers network (TSTNet), that comprises two streams: original and edge. The original stream extracts standard image features whereas the edge stream contains a differentiable edge Sobel operator module and provides edge information. Further, a weighted feature fusion module is introduced to effectively fuse the features from the two streams for boosting the classification performance. The work of \cite{ma2022homo} introduces a transformers-based framework with a  patch generation module designed to generate homogeneous and heterogeneous patches. The patch generation module generates the heterogeneous patches directly, whereas the homogeneous patches are obtained using a superpixel segmentation method.

\begin{figure}[!t]
    \centering
		\includegraphics[width=\linewidth]{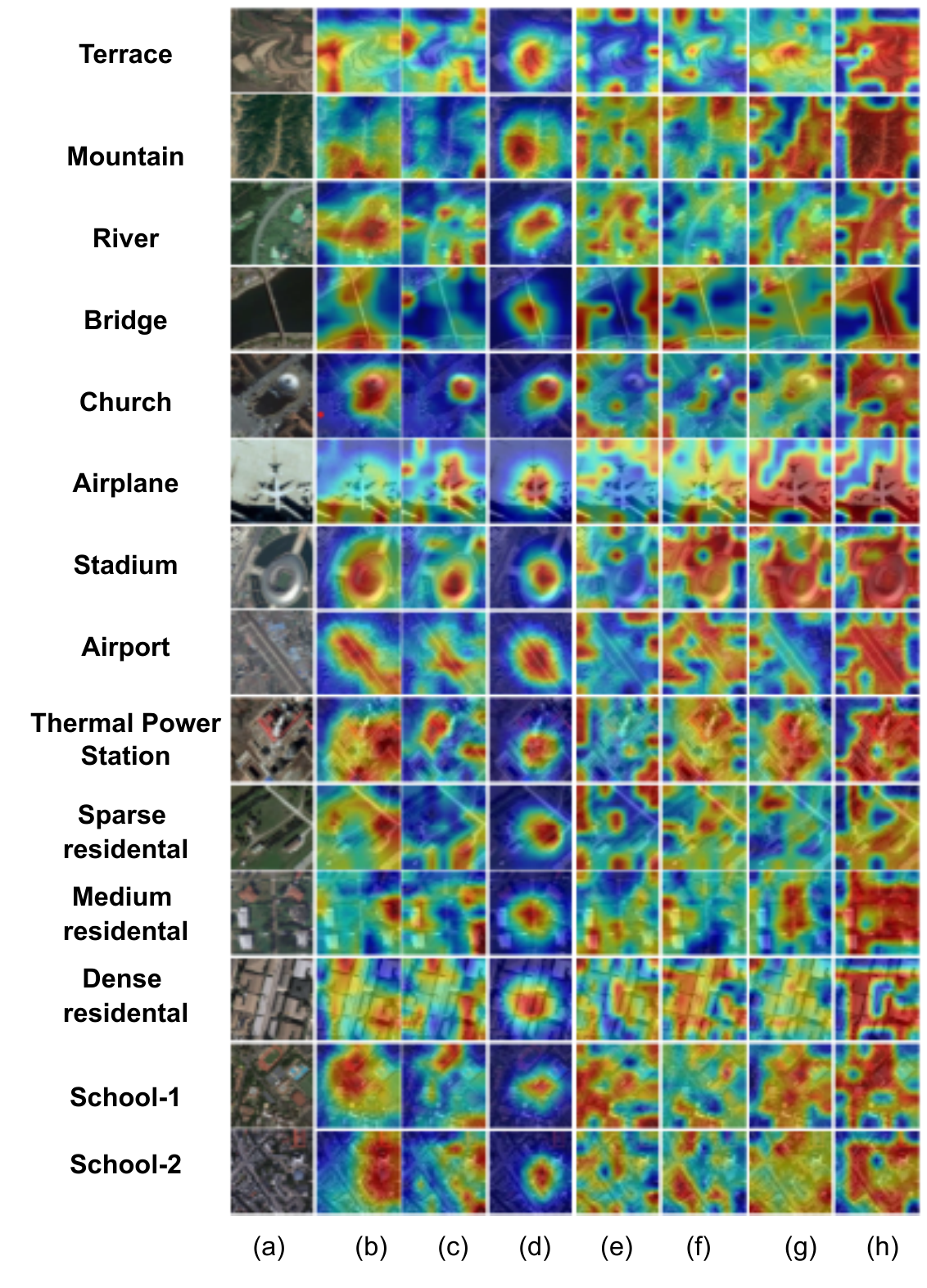}
	\caption{ Comparison in terms of response maps obtained using different models on example VHR images. The original images are shown in (a), whereas the evaluated models are: (b) IMP-ResNet-50, (c) SeCo-ResNet-50, (d) RSP-ResNet-50,
(e) IMP-Swin-T, (f) RSP-Swin-T, (g) IMP-ViTAEv2-S and (h) RSP-ViTAEv2-S. Here, IMP denotes ImageNet pre-training and RSP refers to remote sensing pre-training. In the response map, the warmer color indicates higher response. Figure is from \cite{Wang2022empirical}.}
	\label{fig:Attention_map}
\end{figure}

\textit{Remote Sensing Pre-training:} Different from the aforementioned approaches that either use only transformers or hybrid CNN-transformers designs with backbone networks pretrained on ImageNet datasets, the recent work of
\cite{Wang2022empirical} investigates training vision transformers backbones, such as Swin, from \textit{scratch} on the large-scale MillionAID remote sensing dataset \cite{MillionAID}. The resulting trained backbone models are then fine-tuned for different tasks, including scene classification. Fig.\ref{fig:Attention_map} shows the response maps, obtained using Grad-CAM++ \cite{chattopadhay2018grad}, of different ImageNet (IMP) and remote sensing pre-trained (RSP) models. It can be observed that RSP models learn better semantic representations by paying more attention to the important targets, compared to IMP counterparts. Further, the transformers-based backbones, such as Swin-T better captures the contextual information due to the self-attention mechanism. Moreover, backbones such as, ViTAEv2-S that combines the merits of CNNs and transformers along with RSP can achieve better recognition performance.

Tab.~\ref{tab:clsVHR_tab1} shows a comparison of aforementioned classification approaches on one of the most commonly used VHR classification benchmarks: AID \cite{xia2017aid}. The AID dataset contains images acquired from multi-source sensors. The dataset possesses a high degree of intra-class variation, since the images are collected from different countries, under different time and seasons with variable imaging conditions. There are in total 10,000 images in the dataset and 30 categories. The performance is measured in terms of mean classification accuracy over all the categories. For more details on AID, we refer to \cite{xia2017aid}. Other than RSP that performs an initial pre-training on Million-AID dataset, all approaches here utilize models pre-trained on ImageNet benchmark.

\subsection{Object Detection}\label{MHSA}
 
Localizing objects in VHR imaging is a challenging problem due to extreme scale variations and diversity of different object classes. Here, the task is to simultaneously recognize and localize (either rectangle or oriented bounding-boxes) all instances belonging to different object categories in an image. Most existing approaches employ a hybrid strategy by combining the merits of CNNs and transformers within existing two-stage and single-stage detectors. Other than the hybrid strategy, few recent works also explore DETR-based transformers object detection paradigm \cite{carion2020end}.

\begin{table}[t!]
\scriptsize
\renewcommand{\arraystretch}{1.0}
\begin{center}
\caption{Performance in terms of classification accuracy of different transformers-based methods on the popular AID dataset with 20:80 train-test ratio. }
\label{tab:clsVHR_tab1}
\begin{tabular}{|l|c|c|c|}
\hline
Method      & Venue & Backbone  &  AID (20$\%$)\\
\hline \hline
V16-21K \cite{bazi2021vision}   &   Remote Sensing  &ViT & 94.97\\
CTNet \cite{deng2021cnns}   &  GRSL  & ResNet34 + ViT  & 96.35\\
TRS \cite{zhang2021trs} & Remote Sensing & TRS  & 95.54\\
TSTNet \cite{hao2022two}  & Remote Sensing &  Swin-T  & 97.20\\
RSP \cite{Wang2022empirical}    & TGRS & RSP-Swin-T-E300   & 96.83\\

\hline
\end{tabular}\vspace{-0.5cm}
\end{center}
\end{table}

\textit{Hybrid CNN-Transformers based Methods:} The work of \cite{xu2021improved} introduces a local perception Swin transformer (LPSW) backbone to improve the standard transformers for detecting small-sized objects in VHR imagery. The proposed LPSW strives to combine the merits of transformers and CNNs to improve the local perception capabilities for better detection performance. The proposed approach is evaluated with different detectors such as, Mask RCNN \cite{he2017mask}. The work of \cite{li2022transformer}  introduces  a transformers-based detection architecture, where a pre-trained CNN is used to extract features and a transformer is adapted to process feature pyramid of a remote sensing image.  Zhang \etal \cite{zhang2022gansformer} introduce a detection framework, where an efficient transformer is utilized as a branch network to improve CNN's ability to encode global features. Additionally, a generative model is employed to expand the input remote sensing aerial images ahead of the backbone network. The work of \cite{ADTDet} proposes a detection framework based on RetinaNet, where a feature pyramid
transformer (FPT) is utilized between the backbone network and the post-processing network to generate semantically meaningful features. The FPT enables the interaction among 
features at different levels across scale. The work of \cite{Tangcvpr22} introduces a framework, where transformers are adopted  to model the relationship of
sampled features in order to group them appropriately. Consequently, better grouping and bounding box predictions are obtained without any post-processing operation. The proposed approach effectively eliminates the background information  which helps in achieving improved detection performance.

\begin{figure}[!t]
    \centering
		\includegraphics[width=\linewidth]{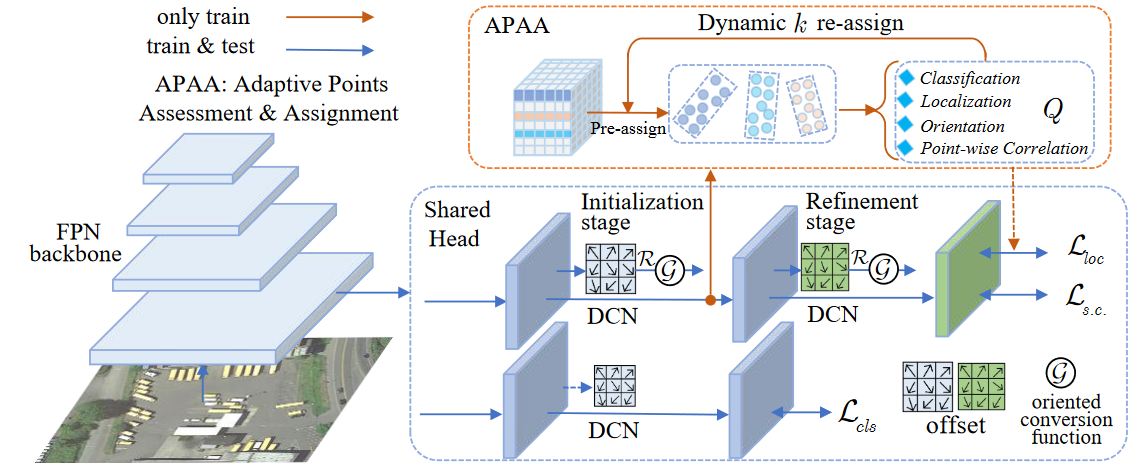}
	\caption{Overview of the  anchor-free Oriented RepPoints detection architecture \cite{li2021oriented} that strives to learn selecting points samples for classification, regression and orientation. RepPoints utilizes same structure of the shared head as in \cite{li2021oriented}, except a quality assessment and sample assignment strategy (APAA) is employed for selecting high-quality sample points for training. Figure is from \cite{li2021oriented}. Best viewed zoomed in. }
	\label{fig:Oriented_RepPoint_VHR}
\end{figure}

Zhang \etal \cite{Rodformer} introduce a hybrid architecture that combines the local characteristics of depth separable convolutions with the global (channel) characteristics of MLP. The work of \cite{PointRCNN}  introduces a two-stage angle-free detector, where both the RPN and regression are angle-free. Their work also evaluates the proposed detector with transformers-based backbone (Swin-Tiny). Liu \etal \cite{HybridNetwork22} propose a hybrid network architecture, called TransConvNet, that aims at combining the advantages of CNNs and transformers by aggregating both global and local information to address the rotation invariability of CNNs with a better contextual attention. Furthermore, an adaptive feature fusion network is designed to capture information from multiple resolutions. The work of \cite{li2021oriented} introduces a detection framework, called Oriented Rep-Points, that utilizes flexible adaptive points as a representation. The proposed anchor-free approach learns to select the point samples from classification, localization and orientation. Specifically, to learn geometric features for arbitrarily-oriented aerial objects, a quality assessment and sample assignment scheme is introduced that measures and identify high-quality sample points for training, as shown in Fig.~\ref{fig:Oriented_RepPoint_VHR}. Further, their approach utilizes a spatial constraint for penalizing the sample points that are outside the oriented box for robust learning of the points. 

\textit{DETR-based Detection Methods:} Few recent approaches have investigated adapting the transformers-based DETR detection framework \cite{carion2020end} for oriented object detection in VHR imaging. The work of \cite{MaOrientedDetr} adapts the standard DETR for oriented object detection. In their approach, efficient encoder is designed for transformers by replacing the standard attention mechanism with a depthwise separable convolution. Dai \etal \cite{AO2DETR} propose a transformers-based detector, called AO2-DETR, where an oriented proposal generation scheme is employed to explicitly produce oriented object proposals. Further, their approach comprises an adaptive oriented proposal refinement module that is designed to compute rotation-invariant features by eliminating the misalignment between region features and objects. Furthermore, a rotation-aware matching loss is utilized to perform a matching process for direct set prediction without the duplicated predictions. 

Tab.\ref{tab:detVHR} shows a comparison of the aforementioned detection approaches on the most commonly used VHR detection benchmark, DOTA \cite{xia2018dota}. The dataset comprises 2,806 large aerial images of 15 different object categories plane, baseball
diamond, basketball court, soccer-ball
field, bridge, ground track field, small
vehicle, Ship, large vehicle,  tennis court, roundabout, swimming pool, harbor, storage tank and helicopter. The detection performance accuracy is measured in terms of mean average precision (mAP). For more details on DOTA, we refer to \cite{xia2018dota}. The results show that most of these recent methods obtain similar detection accuracy, with a slight improvement in performance is obtained when using the Swin-T backbone.

\begin{table}[t!]
\scriptsize
\renewcommand{\arraystretch}{1.0}
\begin{center}
\caption{Comparison in terms of detection accuracy (mAP) of different detectors utilizing a hybrid CNN-transformers design, transformers pre-trained backbone or a DETR-based transformers architecture on DOTA benchmark. The results are presented on the orientated bounding-boxes task of DOTA benchmark. }
\label{tab:detVHR}
\begin{tabular}{|l|c|c|c|}
\hline
Method      & Venue & Backbone  &  DOTA\\
\hline \hline
ADT-Det \cite{ADTDet}    &   Remote Sensing  &ResNet50  & 76.89\\
RBox \cite{Tangcvpr22}    &  CVPR   & ResNet50  & 79.59\\
Rodformer \cite{Rodformer}   &  Sensors  & ResNet50  & 63.89\\
Rodformer \cite{Rodformer}      & Sensors & ViT-B4  & 75.60\\
PointRCNN  \cite{PointRCNN}  & Remote Sensing &  Swin-T  & 80.14\\
Hybrid Network \cite{HybridNetwork22}    & Remote Sensing & TransC-T  & 78.41\\
Oriented RepPoints \cite{li2021oriented}   & Arxiv & ResNet50  & 75.97\\
Oriented RepPoints \cite{li2021oriented}   & Arxiv & Swin-T  & 77.63\\
O$^{2}$DETR \cite{MaOrientedDetr}   & Arxiv & ResNet50  & 79.66\\
AO2-DETR \cite{AO2DETR}    & Arxiv & ResNet50  & 79.22\\
\hline
\end{tabular}\vspace{-0.5cm}
\end{center}
\end{table}

\subsection{Image Change Detection}\label{MHSA}
In remote sensing, image change detection is an important task for detecting changes on the surface of the Earth with numerous applications in agriculture \cite{muzein2006remote,haack1998remote}, urban planning \cite{bolorinos2020consumption}, and map revision \cite{metternicht1999change}. Here, the task is to generate change maps obtained by comparing
the multi-temporal or bi-temporal images, with each pixel in the resulting binary change map having either zero or one value depending on whether the corresponding position has changed or not. Among the recent transformers-based change detection approaches, Chen \etal \cite{ChenCDTGRS} propose a bi-temporal image transformer, encapsulated in a deep feature  differencing-based framework that is designed to model the spatio-temporal contextual information.  Within the proposed framework, the encoder is employed to capture context in token-based space-time. The resulting contextualized tokens are then fed to the decoder where the features are refined in the pixel-space. Guo \etal \cite{guo2021deep} propose a  deep multi-scale Siamese architecture, called MSPSNet, that utilizes a parallel convolutional structure (PCS) and self-attention. The proposed MSPSNet performs feature integration of different temporal images via PCS and then feature refinement based on self-attention to further enhance the multi-scale features.
The work of \cite{zhang2022swinsunet} introduces a Swin transformer-based network with a Siamese U-shaped structure, called SwinSUNet, for change detection. The proposed SwinSUNet comprises three modules:  encoder, fusion, and
decoder. The encoder transforms the input image into tokens and produces multi-scale features by employing a hierarchical Swin transformer. The resulting features are concatenated in the fusion having linear projection and Swin transformer blocks. The decoder contains upsampling and merging within Swin transformer blocks to progressively generate change predictions. 

Wang  \etal  \cite{Guanghui22RS} introduce an architecture, called UVACD, that combines CNNs and transformers for change detection. Within UVACD, the high-level semantic features are extracted via a CNN backbone, whereas transformers are utilized
to generate better change features by capturing the temporal information interaction. The work of \cite{li2022transunetcd} introduces a hybrid architecture, TransUNetCD, that strives to combine the merits of transformers and UNet. Here, the encoder takes features extracted from CNNs and enrich them with global contextual information. The corresponding features are then unsampled and combined with multi-scale features to obtain global-local features for localization. The work of \cite{HybridTransCD} introduces a hybrid multi-scale transformer, called Hybrid-TransCD, that captures both fine-grained and large object features by utilizing heterogeneous tokens via multiple receptive fields.

\begin{table}[t!]
\scriptsize
\renewcommand{\arraystretch}{1.0}
\begin{center}
\caption{Comparison in terms of F1 score of different transformers-based change detection methods on the two popular benchmarks: WHU and LEVIR. }
\label{tab:changeVHR}
\begin{tabular}{|l|c|c|c|}
\hline
Method      & Venue & WHU  &  LEVIR\\
\hline \hline
CD-Trans \cite{ChenCDTGRS}    &   TGRS  &83.98  & 89.31\\
MSPSNet \cite{guo2021deep}    &  TGRS   & -  & 89.18\\
UVACD \cite{Guanghui22RS}   &  Remote Sensing  & 92.84  & 91.30\\
SwinSUNet \cite{zhang2022swinsunet}      & TGRS & 93.8  & -\\
TransUNetCD  \cite{li2022transunetcd}  & TGRS &  93.59  & 91.1\\
HybridTransCD \cite{HybridTransCD}    & IJGI & -  & 90.06\\
\hline
\end{tabular}\vspace{-0.5cm}
\end{center}

\end{table}

\begin{figure*}[t!]
    \centering
		\includegraphics[width=10cm, height=7cm, width=1.0\textwidth]{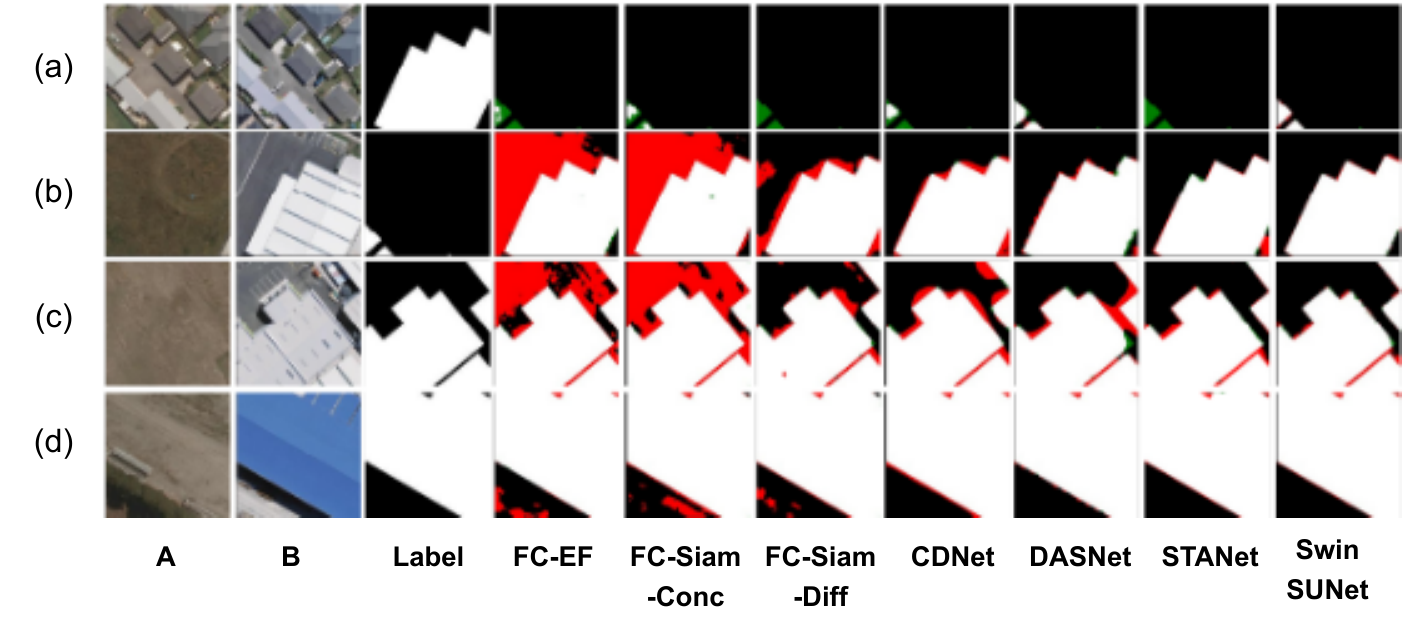}
	\caption{Results of different CD methods visualized, such as FC-EF\cite{daudt2018fully}, FC-Siam-Conc\cite{daudt2018fully}, FC-Siam-Diff\cite{daudt2018fully}, CDNet\cite{alcantarilla2018street}, DASNet\cite{chen2020dasnet}, STANet\cite{chen2020spatial}, and SwinSUNet\cite{zhang2022swinsunet} compared to (a)–(d) sample imagery sets the WHU-CD\cite{ji2018fully} test set. Such as Various colors were utilised to convey different denotations; white represents true positive, black represents true negative, red represents false positive, and green represents false negative. Figure is from \cite{zhang2022swinsunet}.}
	\label{fig:SwinSUNet_Qual}
\end{figure*}

Tab. \ref{tab:changeVHR} shows a comparison of aforementioned change detection approaches on the most commonly used benchmarks: WHU \cite{ji2018fully} and LEVIR \cite{chen2020spatial}. The WHU dataset comprises a single pair of high-resolution (0.075m) images. Here, the images are of size 32507 $\times$ 15354. The LEVIR dataset comprises 637 pairs of high-resolution (0.5m) images. The images are of size 1024 $\times$ 1024. The performance is measured in terms of F1 score with respect to the change category.  Fig. \ref{fig:SwinSUNet_Qual} presents a qualitative comparison of different methods with SwinSUNet on example images from WHU-CD dataset.

\subsection{Image Segmentation}\label{MHSA}

In remote sensing, automatically segmenting an image into semantic categories by performing pixel-level classification is a challenging problem with a wide range of applications, including geological surveys, urban resources management, disaster management and monitoring. Most existing transformers-based remote sensing image segmentation approaches typically employ a hybrid design with an aim to combine the merits of CNNs and transformers. The work of \cite{xu2021efficient} introduces a light-weight transformers-based framework, Efficient-T, that comprises an implicit edge enhancement technique. The proposed Efficient-T employs hierarchical Swin transformers along with MLP head. A coupled CNN-transformers framework, called CCTNet, is introduced in \cite{wang2022cctnet} that aims at combining the local details such as, edges and texture captured by the CNNs along with the global contextual information obtained via transformers for crop segmentation in remote sensing images. Furthermore, different modules such as, test time augmentation and post-processing steps are introduced in order to remove holes and small objects at inference for restoring the complete segmented images. A CNN-transformers framework, named STransFuse, is introduced in \cite{9573374} where both coarse-grained and fine-grained feature representations at multiple scales are extracted and later combined adaptively by utilizing self-attentive mechanism. The work of \cite{9686732} proposes a hybrid architecture, where Swin transformer backbone that captures long-range dependencies is combined with a U-shaped decoder which employs an atrous spatial pyramid pooling block based on depth-wise separable convolution along with SE block to better preserve local details in an image. The work of \cite{panboonyuen2021transformer} utilizes a pre-trained Swin Transformer backbone along with three decoder designs namely, U-Net, feature pyramid network and pyramid scene parsing network for semantic segmentation in aerial images. 

\begin{table}[t!]
\scriptsize
\renewcommand{\arraystretch}{1.0}
\begin{center}
\caption{Performance comparison in terms of overall accuracy (OA) of different transformers-based semantic segmentation methods on two popular benchmarks: Potsdam and Vaihingen. } 
\label{tab:segVHR_}
\begin{tabular}{|l|c|c|c|}
\hline
Method      & Venue & Potsdam  &  Vaihingen\\
\hline \hline
Efficient-T \cite{xu2021efficient}    &   Remote Sensing  &90.08  & 88.41\\
STransFuse \cite{9573374}    &  JSTAR  & 86.71  & 86.07\\
Trans-CNN \cite{9686732}   &  TGRS  & 91.0  & 90.40\\
SwinTF \cite{panboonyuen2021transformer}      & Remote Sensing & -  & 90.97\\
\hline
\end{tabular}\vspace{-0.5cm}
\end{center}

\end{table}

\begin{figure*}[t!]
    \centering
		\includegraphics[width=10cm, height=20cm, width=1.0\textwidth]{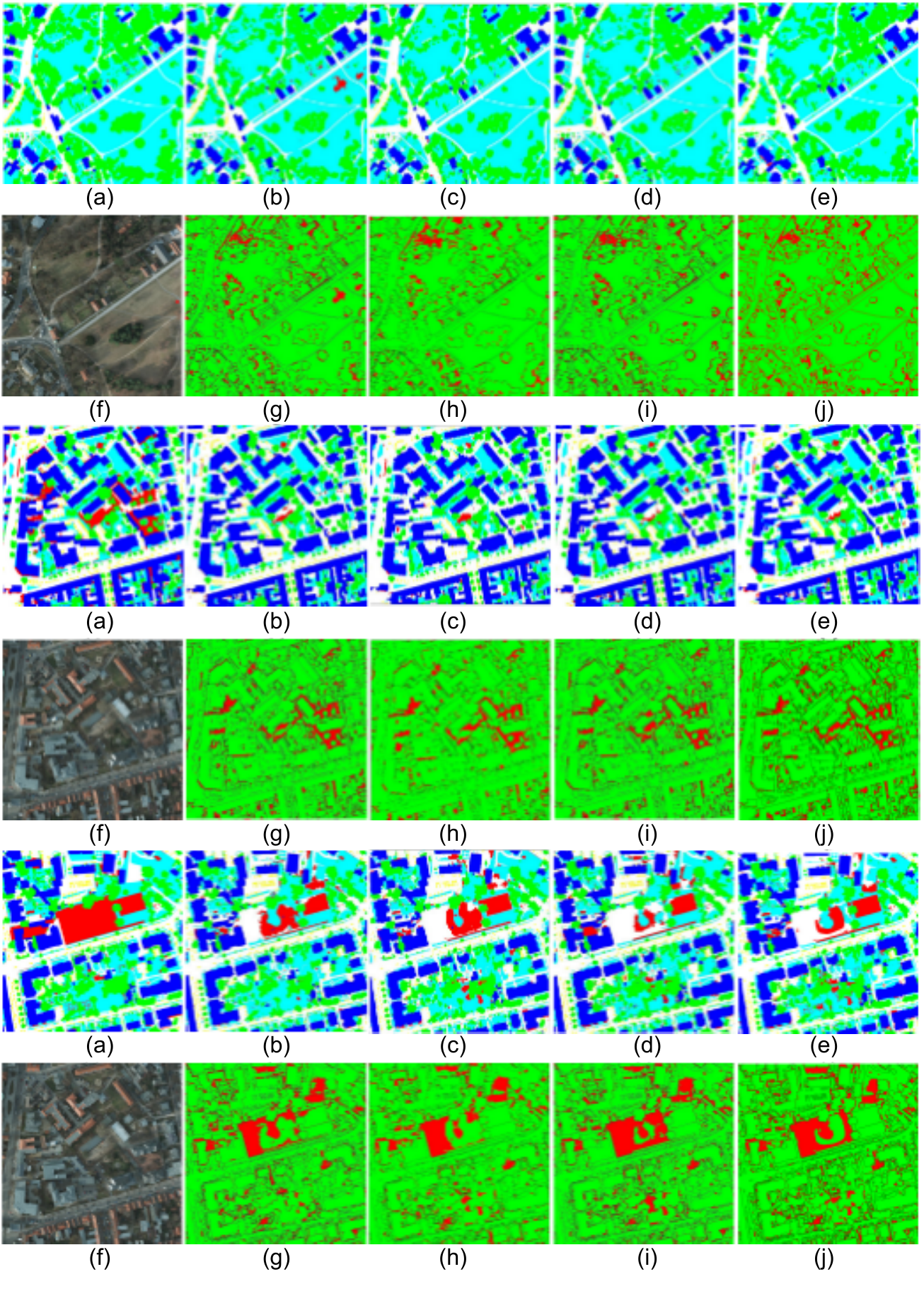}
	\caption{A qualitative comparison between the hybrid Trans-CNN with other existing segmentation approaches. The examples are from the Potsdam dataset. Every  two rows present the results as a group. here, from left to right and top to bottom are: (a) the corresponding ground-truth, (b) results obtained from  AFNet + TTA, (c) results of ResUNet, (d) results of  CASIA2, (e) results achieved using Trans-CNN, and (f) the RGB image. The inccorect classification results from AFNet + TTA,  ResUNet, CASIA2 and Trans-CNN are presented in (g), (h), (i) and (j), respectively. Figure is from \cite{9686732}.}
	\label{fig:VHR_seg_comp}
\end{figure*}

We present in Tab. \ref{tab:segVHR_} a quantitative comparison of aforementioned approaches on the two most commonly used semantic segmentation datasets: Potsdam\cite{Potsdam} and Vaihingen\cite{Vaihingen}. The Potsdam dataset comprises 38 patches, where each patch has a resolution of 6000 $\times$ 6000 pixels collected over the Potsdam City with a ground sampling distance of 5 cm. The dataset has six categories. The Vaihingen dataset comprises 33 samples, where each sample has a resolution from  1996 $\times$ 1995 to 3816 $\times$ 2550 pixels. Here, the ground sampling distance is 9 cm. This dataset contains same categories as Potsdam. The performance is measured in terms of overall accuracy (OA) computed using true positive, false positive, false negative and true negative. Fig.\ref{fig:VHR_seg_comp} presents a qualitative comparison between Trans-CNN and other approaches on the Potsdam dataset.

\textbf{Building Extraction:} Transformers-based techniques have also been recently explored for the problem of building extraction, where the task is to automatically identify building and non-building pixels in a remote sensing image. A dual-pathway transformers framework is introduced in \cite{chen2021building} that strives to learn long-range dependencies both in spatial and channel directions. The work of \cite{xiao2022swin} proposes a transformers framework, STEB-UNet, comprising Swin transformer-based encoding booster that captures semantic information from multi-level features generated from different scales. The encoder booster is further integrated in a  U-shaped network design that fuses local and large-scale semantic features. A transformers-based architectures, called BuildFormer, comprising a window-based linear attention, a convolutional MLP along with batch normalization is introduced in \cite{wang2022buildformer}. The work of \cite{qiu2022transferring} explores the problem of generalizability of building extraction models to different areas and propose a transfer learning approach to fine-tune models from one area
to a subset of another unseen area. 

Other than semantic image segmentation and building extraction with transformers, a recent work by \cite{xu2021improved} explores the problem of instance segmentation where the task is to automatically classify each pixel into an object class within an image while also differentiating multiple object instances. Their approach aims at combining the advantages of CNNs and transformers by designing a local perception Swin transformer backbone to enhance both local and global feature information. 
\subsection{Others}\label{MHSA}
Apart from the problems discussed above, transformers-based techniques are also explored for other VHR remote sensing tasks such as, image captioning and super-resolution. \\ 
\newcolumntype{P}[1]{>{\centering\arraybackslash}p{#1}}

\begin{table*}[h]
\renewcommand{\arraystretch}{1.5}
	\centering
	
	\caption{Overview of transformers-based approaches in VHR remote sensing imaging. Here, we highlight transformers-based methods for different VHR remote sensing tasks.} 	\label{tab:Main_table}
	{\resizebox*{\textwidth}{\textheight}{%
		\begin{tabular}{V{3}l|c|c|c|P{7cm}V{3}} \hlineB{3}
		\rowcolor{mygray} \multicolumn{5}{|c|}{\textbf{Transformers in Very-High Resolution (VHR) Satellite Imagery}} \\ \hlineB{2}
		\rowcolor{mygray}	\textbf{Method}& \textbf{Task}& \textbf{Datasets}& \textbf{Metrics}& \textbf{Highlights} \\\hlineB{2}
			

			 V16-21K \cite{bazi2021vision} &Classification&\makecell{Merced\cite{yang2010bag},\\ AID\cite{xia2017aid},\\ Optimal31\cite{wang2018scene},\\ NWPU\cite{cheng2017remote}}&Overall classification accuracy& Explores vision transformers along with combination of data augmentation techniques for boosting accuracy.\\\hline
			 
			 TRS \cite{zhang2021trs} &Classification&\makecell{Merced\cite{yang2010bag},\\ AID\cite{xia2017aid},\\ Optimal31\cite{wang2018scene},\\ NWPU\cite{cheng2017remote}}&Overall classification accuracy&Integrates transformers into CNNs by replacing the last three ResNet bottlenecks with encoders having multi-head self-attention bottleneck.\\\hline
			 
			 TSTNet\cite{hao2022two}&Classification&\makecell{Merced\cite{yang2010bag},\\ AID\cite{xia2017aid},\\ NWPU\cite{cheng2017remote}}&\makecell{Overall classification accuracy}&A Swin transformer based two-stream architecture that uses both deep features from the image and edge features from edge stream.\\\hline

			 CTNet\cite{deng2021cnns}&Classification &\makecell{AID\cite{xia2017aid},\\ NWPU\cite{cheng2017remote}}&Overall classification accuracy&Comprises a ViT stream that mines semantic features and the CNN stream which captures local structural features.\\\hline

			 HHTL\cite{ma2022homo}&Classification &\makecell{Merced\cite{yang2010bag},\\ AID\cite{xia2017aid},\\ RSSDIVCS\cite{li2021learning},\\ NWPU\cite{cheng2017remote}}&\makecell{Overall classification accuracy}&Explores integrating heterogenous non-overlapping patches and homogenous patches obtained using superpixel segmentation.\\\hline
			 
			 RSP \cite{Wang2022empirical}&Classification, Segmentation, Detection &\makecell{MillionAID\cite{MillionAID},\\ Potsdam \cite{Potsdam},\\ iSAID\cite{waqas2019isaid},\\ HRSC2016 \cite{liu2017high},\\ DOTA\cite{xia2018dota},\\ CCD\cite{lebedev2018change},\\ LEVIR\cite{chen2020spatial}}&\makecell{Overall classification accuracy,\\ mAP,\\ F1 score}&Investigates pre-training transformers on a large-scale remote sensing dataset.\\\hline
			 SAIEC \cite{xu2021improved}&Detection, Segmentation &\makecell{DIOR\cite{li2020object},\\ HRRSD\cite{zhang2019hierarchical},\\ NWPU VHR-10\cite{cheng2016learning}}&\makecell{mAP}&Introduces a local perception Swin transformer backbone that aims to combine the merits of transformers and CNNs for improving the local perception capabilities.\\\hline
			 
			 T-TRD-DA \cite{li2022transformer}&Detection &\makecell{DIOR\cite{li2020object},\\ NWPU VHR-10\cite{cheng2016learning}}
			 &\makecell{mAP}& Proposes a transformers-based detector utilizing a pre-trained CNN for feature extraction and multiple-layer transformers for multi-scale feature aggregation at global spatial positions.  \\\hline
             
			GANsformer \cite{zhang2022gansformer}&Detection &\makecell{DIOR\cite{li2020object},\\ NWPU VHR-10\cite{cheng2016learning}}
			 &\makecell{mAP}& Introduces an efficient transformer, with reduced parameters, as a branch network to capture global features along with a generative model to expand the input image ahead of backbone.\\\hline
			 
			ADT-Det \cite{ADTDet}&Detection &\makecell{DIOR\cite{li2020object},\\ HRSC2016 \cite{liu2017high}}
			&\makecell{mAP}& Introduces a RetineNet-based framework with a feature pyramid transformer integrated between the backbone and post-processing network for generating multi-scale semantic features.\\\hline

            PointRCNN \cite{PointRCNN}&Detection &\makecell{DOTA\cite{xia2018dota},\\ HRSC2016 \cite{liu2017high} }
			&\makecell{mAP}& Introduces a two-stage angle-free dectection framework which is also evaluated using the transformers-based Swin backbone.\\\hline

            HybridNetwork22 \cite{HybridNetwork22}&Detection &\makecell{DOTA\cite{xia2018dota},\\ UCAS-AOD\cite{zhu2015orientation},\\ VEDAI\cite{razakarivony2016vehicle}}&\makecell{mAP}& Integrates multi-scale global and local information from transformers and CNNs through an adaptive feature fusion network.\\\hline
			 
            Oriented RepPoints \cite{li2021oriented}&Detection &\makecell{DOTA\cite{xia2018dota},\\ UCAS-AOD\cite{zhu2015orientation},\\ HRSC2016 \cite{liu2017high}}
			&\makecell{mAP}& Proposes an anchor-free detector learns flexible adaptive points as representations through a quality assessment and sample assignment scheme.\\\hline

            O$^{2}$DETR \cite{MaOrientedDetr}&Detection &\makecell{DOTA\cite{xia2018dota},\\ SKU110K-R\cite{pan2020dynamic},\\ HRSC2016 \cite{liu2017high}}
			&\makecell{mAP}&Extends the standard DETR for oriented detection by introducing an encoder employing depthwise separable convolution. \\\hline

            AO2DETR \cite{AO2DETR}&Detection &\makecell{DOTA\cite{xia2018dota}}
			&\makecell{mAP}&Introduces a DETR-based detector with oriented proposal generation scheme, a refine module to compute rotation-invariant features and a rotation-aware matching loss for performing the matching process for direct set predictions. \\\hline
			
		RBox \cite{Tangcvpr22}&Detection &\makecell{SynthText\cite{gupta2016synthetic},\\ ICDAR 2015 (IC15)\cite{karatzas2015icdar},\\ MLT-2017 (MLT17)\cite{nayef2017icdar2017},\\ MSRA-TD500\cite{yao2012detecting},\\ MTWI\cite{he2018icpr2018},\\ Total-Text\cite{ch2017total},\\ CTW1500\cite{yuliang2017detecting}}
			&\makecell{mAP}&Proposes a framework employing transformers to model the relationship of sampled features for better grouping and box prediction without requiring post-processing operation.\\\hline
			
			Rodformer \cite{Rodformer}&Detection&\makecell{DOTA\cite{xia2018dota}}
			&\makecell{mAP}&A hybrid detection architecture integrating the local characteristics of depth-separable convolutions with the global characteristics of MLP.\\\hline

			CD-Trans \cite{ChenCDTGRS}&Change Detection&\makecell{WHU\cite{ji2018fully},\\ LEVIR\cite{chen2020spatial},\\ DSIFN\cite{zhang2020deeply}}
			&\makecell{F1 score} &Introduces a bi-temporal image transformer designed to model the spatio-temporal contextual information. The encoder captures context in token-based space-time,  which are then fed to decoder where feature refinement is performed in the pixel-space. \\\hline

		\end{tabular}
		}  
}
\end{table*}

\begin{table*}[t!]
 \renewcommand{\arraystretch}{1.5}
 	\centering
 	{\resizebox*{\textwidth}{\textheight}{%
 	\begin{tabular}{V{3}l|c|c|c|P{7cm}V{3}} \hlineB{3}
		
 		\rowcolor{mygray}	\textbf{Method}& \textbf{Task}& \textbf{Datasets}& \textbf{Metrics}& \textbf{Highlights} \\\hlineB{2}

			MSPSNet \cite{guo2021deep}&Change Detection &\makecell{SYSU-CD\cite{shi2021deeply},\\ LEVIR\cite{chen2020spatial}}	&\makecell{F1 score}&Introduces a multi-scale Siamese framework employing a parallel convolutional structure for feature integration of different temporal images and self-attention for feature refinement. \\\hline
			
			SwinSUNet \cite{zhang2022swinsunet}&Change Detection &\makecell{CCD\cite{lebedev2018change},\\ WHU\cite{ji2018fully},\\ OSCD\cite{daudt2018urban},\\ HRSCD\cite{daudt2019multitask}}
			&\makecell{F1 score}&Introduces a Swin transformer-based network with a Siamese U-shaped structure having encoder, fusion and decoder modules.\\\hline
			
            TransUNetCD \cite{li2022transunetcd}&Change Detection &\makecell{WHU\cite{ji2018fully},\\ LEVIR\cite{chen2020spatial},\\ CCD\cite{lebedev2018change},\\ DSIFN\cite{zhang2020deeply}, \\OSCD\cite{daudt2018urban},\\ S2Looking\cite{shen2021s2looking}}
			&\makecell{F1 score}&Introduces a framework integrating merits of transformers and UNet through capturing enriched contextualized features which are upsampled and fused with multi-scale features to generate global-local features. \\\hline
			
			Hybrid-TransCD \cite{HybridTransCD}&Change Detection &\makecell{LEVIR\cite{chen2020spatial},\\ SYSU-CD\cite{shi2021deeply}}
			&\makecell{F1 score}&Introduces a multi-scale transformer that encodes both fine-grained and large object features through heterogeneous tokens via multiple receptive fields.\\\hline

			 CCTNet \cite{wang2022cctnet}&Segmentation &\makecell{Barley Remote Sensing Dataset \cite{barley}}&\makecell{F1 score,\\ Overall accuracy}& Proposes a hybrid CNN-transformers framework to combine local details and global conextual information for crop segmentation. \\\hline
			 
			 STransFuse \cite{9573374}&Segmentation &\makecell{Potsdam\cite{Potsdam},\\ Vaihingen\cite{Vaihingen}}&\makecell{F1 score,\\ Overall accuracy}& Introduces a framework that encodes both coarse-grained as well as fine-grained features at multiple scales which are fused using self-attentive mechanism. \\\hline
			 
			 Trans-CNN \cite{9686732}&Segmentation&\makecell{Potsdam\cite{Potsdam},\\ Vaihingen\cite{Vaihingen}}&\makecell{F1 score,\\ Overall accuracy}& Introduces a framework with a Swin transformer backbone to capture long-range dependencies and a U-shaped decoder with depth-wise separable convolution to encode local details. \\\hline
			 
			 
		SwinTF \cite{panboonyuen2021transformer}&Segmentation &\makecell{Vaihingen\cite{Vaihingen},\\ Thailand North Landsat-8 corpus (private),\\ Thailand Isan Landsat-8 corpus (private)}&\makecell{F1 score,\\ Overall accuracy}&Introduces a framework with pre-trained Swin backbone along with a U-Net, feature pyramid network and a pyramid scene parsing network for segmentation. 
\\\hline
			 
			Efficient-T \cite{xu2021efficient}&Segmentation &\makecell{Potsdam\cite{Potsdam},\\ Vaihingen\cite{Vaihingen} }&\makecell{F1 score,\\ Overall accuracy}& Proposes a light-weight framework consisting of an implicit edge enhancement scheme along with a Swin transformers.\\\hline

            STT \cite{chen2021building}&Building Extraction &\makecell{WHU\cite{ji2018fully},\\ INRIA\cite{maggiori2017can}}&\makecell{IoU,\\ Overall accuracy,\\ F1 score}&Introduces a transformers framework to learn long-range dependencies both in the spatial and channel direction. \\\hline
			 
			 STEB-UNet \cite{xiao2022swin}&Building Extraction &\makecell{WHU\cite{ji2018fully},\\ Massachusetts\cite{maggiori2017can}}&\makecell{IoU,\\ F1 score}&Introduces a transformers framework capturing semantic information from multi-scale features which are further fused to local features. \\\hline
			 
			 BuildFormer \cite{wang2022buildformer}&Building Extraction &\makecell{WHU\cite{ji2018fully},\\ Massachusetts\cite{maggiori2017can},\\
			 INRIA\cite{maggiori2017can}}& \makecell{IoU,\\ F1 score}& Introduces an architecture consisitng of a window-based linear attention and a convolutional MLP.\\\hline
			 
			 T-Trans \cite{qiu2022transferring}&Building Extraction &\makecell{Massachusetts\cite{maggiori2017can}\\
			 ,INRIA\cite{maggiori2017can}}&\makecell{IoU,\\ F1 score}& Explores the task of generalizability of building extraction models to different areas and introduces a transfer learning method to fine-tune models from one area to a subset of another unseen area.  \\\hline
			 
 
		    TRL \cite{shen2020remote}&\makecell{Image Captioning} &\makecell{RSICD\cite{lu2017exploring},\\ UCM-captions \cite{UCMCaptions},\\ Sydney-Caption\cite{sydney}}
			&\makecell{BLEU,\\ ROUGE,\\ METEOR,\\ CIDEr}&Proposes an approach adapting transformers by integrating residual connections, dropout and adatpive feature fusion for remote sensing image caption generation. \\\hline
			
			MLAT\cite{liu2022remote}&\makecell{Image Captioning} &\makecell{RSICD\cite{lu2017exploring},\\ UCM-captions\cite{UCMCaptions},\\ Sydney-Caption\cite{sydney}]}
			&\makecell{BLEU,\\ ROUGE,\\ METEOR,\\ CIDEr}&Introduces an architecture where multi-scale features from CNN layers are extracted in encoder and a multi-layer aggregated transformer in the decoder uses those features for sentence generation. \\\hline
			
			Ren \etal\cite{ren2022mask}&\makecell{Image Captioning} &\makecell{RSICD\cite{lu2017exploring},\\ UCM-captions\cite{UCMCaptions},\\ Sydney-Caption\cite{sydney}}
			&\makecell{BLEU,\\ ROUGE,\\ METEOR,\\ CIDEr}&Proposes a topic token-based mask transformers  with the topic token being integrated into encoder while serving as prior in decoder for capturing global semantic relationships.\\\hline
			

			TR-MISR \cite{an2022tr}&\makecell{Image Super Resolution}&\makecell{RSICD\cite{lu2017exploring},\\ UCM-captions\cite{UCMCaptions},\\ PROBA-V\cite{martens2019super}}
			&\makecell{cPSNR,\\ cSSIM}&Introduces a transformers-based architecture with an encoder having residual blocks, a fusion module along with a super-pixel convolution-based decoder for multi-image super-resolution. \\\hline
			
			MSE-Net \cite{lei2021transformer}&\makecell{Image Super Resolution}&\makecell{UCMerced\cite{ucmerced},\\ AID\cite{xia2017aid}}
			&\makecell{cPSNR,\\ cSSIM}&Proposes a multi-stage enchancement framework to utilize features from different stages and further integrating them with standard super-resolution technique for combining multi-resolution low as well as high-dimension feature representations. \\\hline
			
		SRT \cite{ye2021super}&\makecell{Image Super Resolution}&\makecell{UCMerced\cite{ucmerced}}
			&\makecell{cPSNR,\\ cSSIM}&Introduces a hybrid framework that integrates local features from CNNs and global features from transformers. \\\hline

 		\end{tabular}}
 		
 	}
 	
\end{table*}
\afterpage{\clearpage}
\textbf{Image Captioning:}
Image captioning in remote sensing images is a challenging problem, where the task is to generate semantically natural description of a given image. Few recent works have explored using transformers for image captioning. The work of \cite{shen2020remote} introduces a framework, where standard transformers are adapted for remote sensing image caption generation by integrating  residual connections, dropout layers and fusing features adaptively. Moreover, a reinforcement learning technique is utilized to further improve the caption generation process. An encoder-decoder architecture is introduced in \cite{liu2022remote}, where the multi-scale features are first extracted from different layers of CNNs in the encoder and then a multi-layer aggregated transformer is utilized in the decoder to effectively exploit the multi-scale features for generating sentences. The work of \cite{ren2022mask} introduces a topic token-based mask transformers framework, where a topic token is integrated into the encoder and serves as a prior in the decoder for capturing improved global semantic relationships.\\ 
\textbf{Image Super Resolution}
Remote sensing image super-resolution is the task of recovering high-resolution images from their low-resolution counterparts. A few recent works have explored transformers for this task. A transformer-based multi-stage enhancement structure is introduced in \cite{lei2021transformer} that leverages features from different stages. The proposed multi-stage structure can be combined with conventional super-resolution techniques in order to fuse multi-resolution low as well as high-dimension features. \cite{ye2021super} proposes a CNN-transformer hybrid architecture to integrate both local and global feature information for super-resolution. The work of \cite{an2022tr} explores the problem of multi-image super-resolution, where the task is to merge multiple low-resolution remote sensing images of the same scene into a high-resolution one. Here, a transformers-based approach is introduced comprising an encoder having residual blocks, a fusion module and a super-pixel convolution-based decoder. 

To summarize the review of transformers in VHR imagery, we present a holistic overview of different techniques in literature in Tab.\ref{tab:Main_table}.

\section{Transformers in Hyperspectral Imaging}\label{sec:HIA}

As discussed earlier, hyperspectral images  are represented by several spectral brands and analyzing hyperspectral data is crucial in a wide range of problems. Here, we present a review of recent transformers-based approaches for different hyperspectral imaging (HSI) tasks. 

\subsection{Image Classification}\label{MHSA}

Here, the task is to automatically classify and assign a category label to each pixel in an image acquired through hyperspectral sensors. Next, we first review recent works that are either based on the pure transformers design or utilize a hybrid CNN-transformers approach. Afterwards, we discuss few recent transformers-based approaches fusing different modalities for hyperspectral image classification.

\textit{Pure Transformers-based Methods:} Among existing works, the approach of \cite{he2019hsi} introduces a bi-directional encoder representation from transformers, called HSI-BERT, that strives to capture global dependencies. The proposed architecture is flexible and can be generalized from different regions with the need to perform pre-training. A transformers-based backbone, called  SpectralFormer, is introduced in \cite{hong2021spectralformer}  that can take pixel-wise or patch-wise inputs and  is designed to capture spectrally local sequence knowledge from nearby hyperspectral bands. SpectralFormer utilizes  cross-layer skip connection to circulate information from shallow to deep layers by learning soft residuals across layers, thereby producing  
group-wise spectral embeddings. To circumvent the problem of fixed geometric structure of convolution kernels, a spectral–spatial transformer network is proposed in \cite{zhong2021spectral} comprising a spatial attention and a spectral association module. While the spatial attention aims at connecting the local regions through aggregation of all input feature channels with spatial kernel weights, the spectral association is achieved through the integration of all spatial locations of the corresponding masked feature maps. Transformers are also explored in the spatial and spectral dimensions in \cite{liu2022dss}. Here, a framework is introduced comprising spectral self-attention that learns to capture interactions along the spectral dimension and a spatial self-attention designed to pay attention to features along the spatial dimension. The resulting features from both spectral and spatial self-attention are then combined and input to the classifier. 

\begin{figure}[!t]
    \centering
		\includegraphics[width=\linewidth]{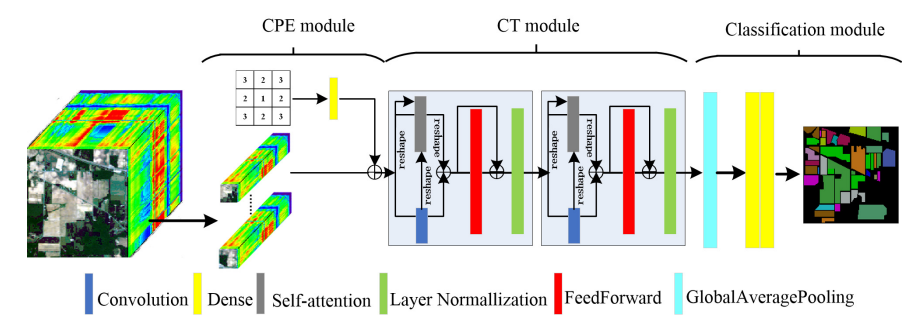}
	\caption{Overview of the CTN framework \cite{zhao2022convolutional} for hyperspectral image classification. Given the HSI data patches, CTN processes them to  center position encoding (CPE), convolutional transformer 
and classification modules. Here, the output represents the category label. Figure is from \cite{zhao2022convolutional}. Best viewed zoomed in.}
	\label{fig:HSI_CTN_Method}
\end{figure}

\textit{Hybrid CNN-Transformers based Methods:} Several works recently have explored combining the merits of CNNs and transformers to better capture both the local information as well as long-range dependencies for hyperspectral image classification. To this end, a convolutional transformer network, named CTN, is introduced in \cite{zhao2022convolutional} that utilizes center position encoding to generate spatial position features by combining pixel positions with spectral features as well as convolutional transformer to further obtain local-global features, as shown in Fig.\ref{fig:HSI_CTN_Method}. A  hyperspectral image transformer (HiT) classification approach is proposed in \cite{yang2022hyperspectral}, where convolutions are embedded into transformers architecture to further integrate local spatial contextual information. The proposed approach comprises two main modules, where one module, called  spectral-adaptive 3-D convolution projection, is designed to generate spatial-spectral local information via spectral adaptive 3D convolution layers from hyperspectral images. The other module, named Conv-Permutator,  employs depthwise convolutions to capture spatial–spectral representations separately along the spectral, height and width dimensions. 
The work of \cite{jia2022multiscale} introduces a multi-scale convolutional transformer that effectively captures spatial-spectral information which can be integrated with transformers network. Further, a self-supervised pre-task is defined that masks the token of the central pixel in the encoder, whereas remaining tokens are input to the decoder in order to reconstruct the spectral information corresponding to the central pixel. In \cite{sun2022spectral}, a spectral–spatial
feature tokenization transformer, called SSFTT, is proposed that generates spectral-spatial and semantic features. The SSFTT comprises a  feature extraction module that produces low-level spectral and spatial features by employing a 3D and a 2D convolution layer. Furthermore, a  Gaussian weighted feature tokenizer is utilized in SSFTT for feature transformation which are then input to a transformer encoder for feature
representation. Consequently, a linear layer is employed to generate the sample label. Zhao \textit{et al.} \cite{zhao2022convolutional} 
proposes a convolutional transformer network (CTN) that employs center position encoding to combine spectral features with pixel positions. The proposed architecture introduces convolutional transformer blocks that effectively integrates local and global features from hyperspectral image patches. Yang \textit{et al.} \cite{yang2022hyperspectral} introduces a hyperspectral image transformer (HiT) framework, where convolution operations are embedded within the transformers design for also integrating local spatial contextual information. The HiT framework comprises of a spectral-adaptive 3D convolution projection to capture local spatial-spectral information. Additionally, the HiT framework employs a  conv-permutator module that uses the depthwise convolution for explicitly capturing the spatial-spectral information along different dimensions: height, width and spectral. The work of \cite{sun2022spectral} introduces a spectral–spatial feature tokenization transformer, named SSFTT, that consists of a spectral-spatial feature extraction scheme for encoding shallow spectral-spatial features, a feature transformation module which produces transformed features used as input in the encoder.

\textit{Multi-modal Fusion Transformers based Methods:}
Few recent transformers-based works also explore fusing different modalities, such as hyperspectral, SAR, LiDAR for hyperspectral image classification. A Multi-modal fusion transformer, MFT, is introduced in \cite{roy2022multimodal}, that comprises a data fusion scheme to derive class tokens in the transformers from multi-modal data (e.g., LiDAR, SAR) along with the standard hyperspectral patch tokens. Further, the attention mechanism within MFT fuses information from tokens of hyperspectral and other modalities into a new token of integrated features. The work of \cite{xue2022deep} introduces an approach, where a spectral sequence transformer is utilized to extract features from hyperspectral images along the spectral dimension and a spatial hierarchical transformer to generate spatial features in a hierarchical manner from both hyperspectral and LiDAR data. 
\begin{figure*}[t!]
    \centering
		\includegraphics[width=1.0\textwidth]{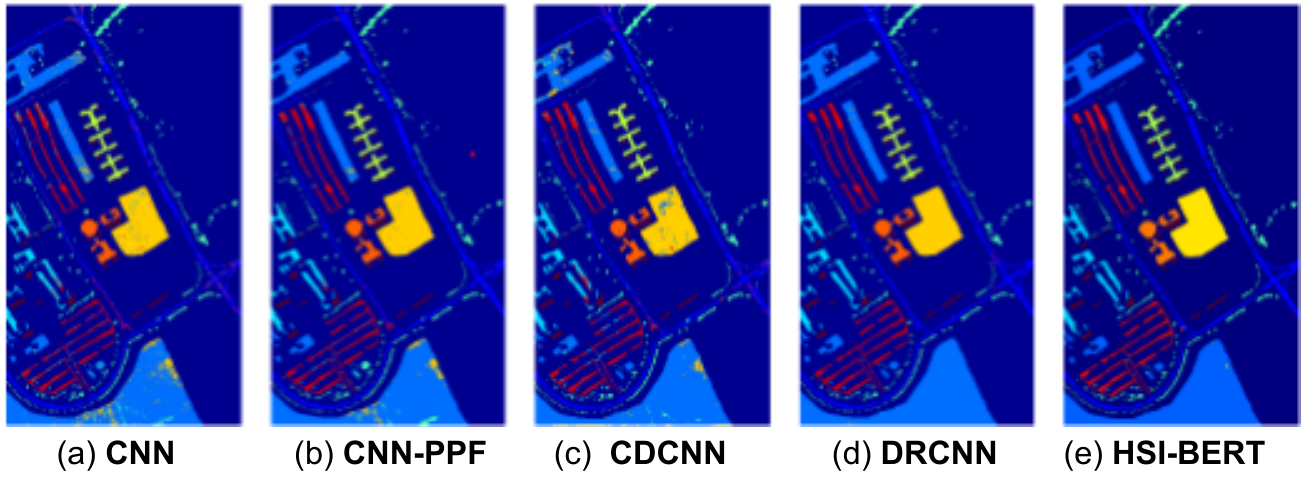}
	\caption{A qualitative comparison, in terms of visualization of classification maps, between HSI-BERT and several CNN-based methods on the Pavia dataset. Here, (a) CNN, (b) CNN-PPF, (c) CDCNN, (d) DRCNN, and (e) HSI-BERT. Figure is from \cite{he2019hsi}.}
	\label{fig:HSIBert_CLs_comp}
\end{figure*}

\begin{table}[t!]
\scriptsize
\renewcommand{\arraystretch}{1.0}
\begin{center}
\caption{Comparison in terms of overall accuracy (OA) of some representative CNN-based methods with pure transformers and hybrid CNN-transformers based hyperspectral image classification methods on two popular benchmarks: Indian Pines and Pavia. Here, the results are reported using 200 samples for training for each category.  }
\label{tab:segHSI}

\begin{tabular}{|l|c|c|c|c|}
\hline
Method      & Venue & Type & Indian Pines &  Pavia  \\
\hline \hline
CNN \cite{HuSensors}   &  Sensors  & CNNs  & 87.01& 92.27\\
CNN-PPF \cite{Li2018fully}   &  TGRS  & CNNs  & 93.90& 96.48\\ \hline
HSI-BERT \cite{he2019hsi}    &   TGRS & Pure&   99.56  & 99.75 \\
 DSS-TRM \cite{liu2022dss}    &  EJRS  & Pure  & 99.43& 98.50\\
CTN \cite{zhao2022convolutional}   &  GRSL  & Hybrid  & 99.11& 97.48\\
\hline
\end{tabular}\vspace{-0.5cm}
\end{center}

\end{table}

Tab.\ref{tab:segHSI} shows a comparison of some representative CNN-based approaches with both pure transformers and hybrid CNN-transformers based methods on two popular hyperspectral image classification benchmarks: Indian Pines and Pavia. The Indian Pines dataset is acquired through airborne visible/infrared imaging spectrometer (AVIRIS) sensor in Northwestern Indiana, USA. Here, the images comprise 145 $\times$ 145 pixels  in the spatial dimension, at a ground sampling distance (GSD) of 20m with 220 spectral bands that cover the wavelength range of 400–2500 nm. After the removal of noisy bands, 200 spectral brands are retained. The original dataset contains 16 class, where several methods discard the small classes. For the remaining categories, the number of training samples are 200 per class. The Pavia dataset comprises images acquired through the reflective optics system imaging spectrometer (ROSIS) sensor over Pavia, Italy. Here, the images consist of 610 $\times$ 340 pixels  in the spatial dimension, at a GSD of 1.3m with 103 spectral bands covering from 430 to 860 nm. The dataset contains nine categories, where the number of training samples are 200 per class. Generally, three metrics are used to evaluate the performance of methods quantitatively: overall accuracy, average
accuracy and kappa coefficient. The overall accuracy (OA) denotes to the proportion of correctly classified test samples, whereas average accuracy (AA) reflects the average recognition accuracy for each category. The kappa coefficient refers to the consistency between the generated classification maps from the model and the available ground-truth. Fig.\ref{fig:HSIBert_CLs_comp} presents a qualitative comparison between HSI-Bert \cite{he2019hsi} and other existing CNN-based methods on the Pavia dataset.



\begin{table*}
	\centering
		\caption{Overview of transformers-based approaches in Hyperspectral imaging. Here, we highlight methods for different Hyperspectral remote sensing tasks.} \label{tab:HSISAR}
	\resizebox{\textwidth}{!}{
	\begin{tabular}{V{3}l|c|c|c|P{7cm}V{3}} \hlineB{3}
		\rowcolor{mygray} \multicolumn{5}{|c|}{\textbf{Transformers in Hyperspectral Imagery}} \\ \hlineB{2}
		\rowcolor{mygray}	\textbf{Method}& \textbf{Task}& \textbf{Datasets}& \textbf{Metrics}& \textbf{Highlights} \\\hlineB{2}
			

			 SpectralFormer \cite{hong2021spectralformer} & Classification&\makecell{Indian Pines\cite{indian},\\ Pavia University\cite{pavia},\\ {Houston2013}\cite{houston}}&\makecell{Overall classification accuracy,\\ Kappa}&  Introduces a transformers-based backbone to capture spectrally local information from nearby hyperspectral bands by generating group-wise spectral embeddings.
\\\hline
			 
			 MCT \cite{jia2022multiscale} &Classification&\makecell{Salinas\cite{salinas},\\ Yellow River Estuary}&\makecell{Overall classification accuracy,\\ Kappa}& Proposes a multi-scale convolutional transformer to encode spatial-spectral information that is integrated with transformers network.\\\hline
			 
			 MFT\cite{roy2022multimodal}&Classification&\makecell{University of Houston\cite{houston},\\ Trento,\\ MUUFL Gulfport\cite{gader2013muufl},\\ Augsburg scenes}&\makecell{Overall classification accuracy,\\ Kappa}& Proposes a multi-modal transfomers that derives class tokens from multi-modal data along with the standard hyperspectral patch tokens.
\\\hline
 
 			 CTN \cite{zhao2022convolutional} &Classification&\makecell{Indian Pines\cite{indian},\\ Pavia University\cite{pavia}}&\makecell{Overall classification accuracy,\\ Kappa}&  Introduces a convolutional transformer network with dedicated blocks that integrates local and global features from hyspectral image patches.\\\hline
			 
			 DHViT \cite{xue2022deep} &Classification&\makecell{Trento,\\ Houston 2013\cite{houston},\\ Houston 2018\cite{Houston18}}&\makecell{Overall classification accuracy,\\ Kappa}& Introduces an approach comprising a spectral sequence transformer to encode features along the spectral dimension and a spatial hierarchical transformer to produce hierarchical spatial features for hyperspectral and LiDAR data. \\\hline
			 
			 DSS-TRM \cite{liu2022dss}&Classification&\makecell{Pavia University\cite{pavia},\\ Salinas\cite{salinas},\\ Indian Pines\cite{indian}}&\makecell{Overall classification accuracy,\\ Kappa}& Introduces a transformers-based approach consisting of spectral self-attention  and spatial self-attention to capture interactions along spectral and spatial dimension, respectively. \\\hline
			 
 			 HiT \cite{yang2022hyperspectral}&Classification&\makecell{Indian Pines\cite{indian},\\ Pavia University\cite{pavia},\\ Houston2013\cite{houston},\\ Xiongan}&\makecell{Overall classification accuracy, Kappa}& Proposes a hyperspectral image transformer consisting of a 3D convolution projection module to encode local spatial-spectral details and a conv-permutator modue to capture the information along height, width and spectral dimensions. \\\hline
 
 			 HSI-BERT \cite{he2019hsi} &Classification&\makecell{Indian Pines\cite{indian},\\ Pavia University\cite{pavia},\\ Salinas\cite{salinas}}&\makecell{Overall classification accuracy}&  Proposes a transformers-based method that captures capture global dependencies using a bi-direction encoder representation. \\\hline
			 
			 SSFTT \cite{sun2022spectral} &Classification&\makecell{Indian Pines\cite{indian},\\ Pavia University\cite{pavia},\\ Houston 2013\cite{houston}}&\makecell{Overall classification accuracy,\\ Kappa}& Proposes a spectral–spatial feature tokenization transformer that utilizes both spectral-spatial shallow and semantic features for representation and learning. \\\hline
			 
			 SSTN \cite{zhong2021spectral}&Classification&\makecell{Pavia University\cite{pavia},\\ Kennedy Space Center,\\ Indian Pines\cite{indian},\\ University of Houston\cite{houston},\\ Pavia Center\cite{paviacenter}}&\makecell{Overall classification accuracy,\\ Kappa}& Introduces a spectral–spatial transformer with a spatial attention and a spectral association module. The two modules perform spectral and spatial association through the integration of spectral and spatial locations, respectively. \\\hline

			 CTIN \cite{zhou2022panformer}&Pan-Sharpening&\makecell{worldview II\cite{WVii},\\ worldview III\cite{wviii},\\ GaoFen-2}&\makecell{IQA,\\ ERGAS,\\ PSNR,\\ SAM}& A transformers-based approach is introduced, where multi-spectral and panchromatic features are captured for joint feature learning across modalities. Further, an invertible neural module performs feature fusion to generate pansharpened images. \\\hline
			 
			 HyperTransformer \cite{Bandaracvpr22}&Pan-Sharpening&\makecell{Pavia Center\cite{paviacenter},\\ Botswana\cite{Botswana},\\ Chikusei\cite{NYokoya2016}}&\makecell{Cross-correlation(CC),\\ Spectral Angle Mapping (SAM),\\ RSNR,\\ ERGAS,\\ PSNR}& Introduces a transformers-based framework with separate feature extractors for panchromatic and hyperspectral images and a spectral-spatial fusion module to learn cross-feature space dependencies of features. \\\hline
			 
			 PMACNet \cite{liang2022pmacnet}&Pan-Sharpening&\makecell{worldview II\cite{WVii},\\ worldview III\cite{wviii}}&\makecell{spatial correlation coefficient(SCC),\\ spectral angle mapper (SAM)}& Introduces a framework with a parallel CNN structure to learn ROIs from low-resolution image and residuals from high-resolution image. It also contains a a pixel-wise attention module to adapt residuals on the learned ROIs.\\\hline			 
			 
			 CPT-noRef \cite{li2022pan}&Pan-Sharpening&\makecell{Gaofen-1,\\ worldview II\cite{WVii},\\ Pleiades\cite{pleiades}}&\makecell{IQA,\\ ERGAS,\\ SAM,\\ correlation coefficient(CC)}& A CNN-transformers framework where global features are generated using transformers and local features are constructed using a shallow CNNs. The features are combined and a loss formulation having spatial and spectral losses are utilized for training.\\\hline
			 
			 MSIT \cite{zhang2022multiscale}&Pan-Sharpening&\makecell{GeoEye-1,\\ QuickBird\cite{QB}}&\makecell{ERGAS,\\ SAM,\\ Q4}& Introduces a multi-scale spatial–spectral interaction transformer with a convolution-transformer encoder for generating multi-scale global and local features from both low-resolution and panchromatic images.\\\hline			 
			 
			 Su \etal \cite{su2022transformer}&Pan-Sharpening&\makecell{worldview II\cite{WVii}, QuickBird\cite{QB}, GaoFen-2}&\makecell{spatial correlation coefficient(SCC),\\ ESGAS,\\ RMSE,\\ SAM,\\ Q4}& A transformers-based approach with spatial and spectral feature extraction performed using a Swin model. \\\hline

 		\end{tabular}
 		}

\end{table*}
\subsection{Hyperspectral Pansharpening}\label{MHSA}
In the hyperspectral pansharpening problem, the task is to enhance low-resolution hyperspectral image spatially using the spatial information from registered panchromatic image, while preserving the spectral information of the low-resolution image. Pansharpening plays an important role in a variety of tasks in remote sensing, including classification and change detection. Previously, CNN-based approaches have shown promising results for this task. Recently, transformers-based methods have performed favorably for this problem by also utilizing the useful global contextual information. A multi-scale spatial–spectral interaction transformer, MSIT, is proposed by \cite{zhang2022multiscale} that comprises a convolution–transformer encoder to extract multi-scale local and global features from low-resolution and panchromatic images. The work of \cite{li2022pan} introduces an architecture, where global features are constructed using  transformers and local features are computed using a shallow CNN. These multi-scale features extracted in a pyramidal fashion are learned simultaneously. The proposed approach further introduces a loss formulation with spatial and spectral loss simultaneously used for training using the real data. Liang \textit{et al.} \cite{liang2022pmacnet} propose a framework, named PMACNet, where both the region-of-interest from the low-resolution image and the residuals for  regression to high-resolution image are learned in a parallel CNN structure. Afterwards, a pixel-wise attention module is utilized to adapt the residuals based on the learned region-of-interest.

A transformers-based regression network is introduced by \cite{su2022transformer}, where the feature extraction of spatial and spectral information is performed by utilizing a Swin transformer model. The work of  \cite{DBLP:conf/aaai/ZhouHFFL22} introduces a transformers-based approach, where  multi-spectral and panchromatic features are formulated as keys and queries for enabling joint learning of features across the modalities. Further, this work employs an invertible neural module to perform effective fusion of the features for generating the pansharpened images. Bandara \textit{et al.} \cite{Bandaracvpr22} propose a framework comprising separate feature extractors for panchromatic and hyperspectral images, a soft attention mechanism and a spectral-spatial fusion module. The pansharpened image quality is improved by learning cross-feature space dependencies of the different features.

To summarize the review of transformers in hyperspectral imaging, we provide a holistic overview of the existing techniques in literature in Tab.\ref{tab:HSISAR}.

\section{Transformers in SAR Imagery}\label{sec:SARS}
As discussed earlier, SAR images are constructed from the signals of the electromagnetic waves, through a a sensor platform, transmitted to the surface of Earth. SAR possesses unique characteristics due to being unaffected with different environmental conditions such as, day, night and fog. Here, we review recent transformers-based approaches for SAR imaging tasks.


\subsection{SAR Image Interpretation}\label{MHSA}
\begin{figure}[!t]
    \centering
		\includegraphics[width=\linewidth]{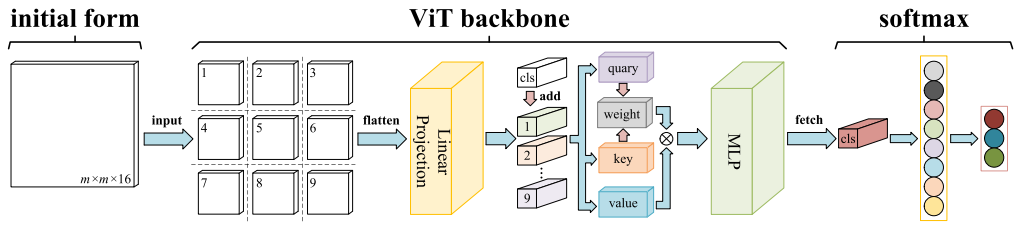}
	\caption{Overview of the ViT-PolSAR framework \cite{9658539} for supervised polarimetric
SAR image classification. Here, the pixel values of the SAR image patches are considered as tokens and then the self-attention mechanism is utilized to encode longe-range dependencies followed by MLP. Figure is from \cite{9658539}. Best viewed zoomed in. }
	\label{fig:SAR_PoISAR_Method}
\end{figure}

\begin{figure*}[t!]
    \centering
		\includegraphics[width=10cm, height=7cm, width=1.0\textwidth]{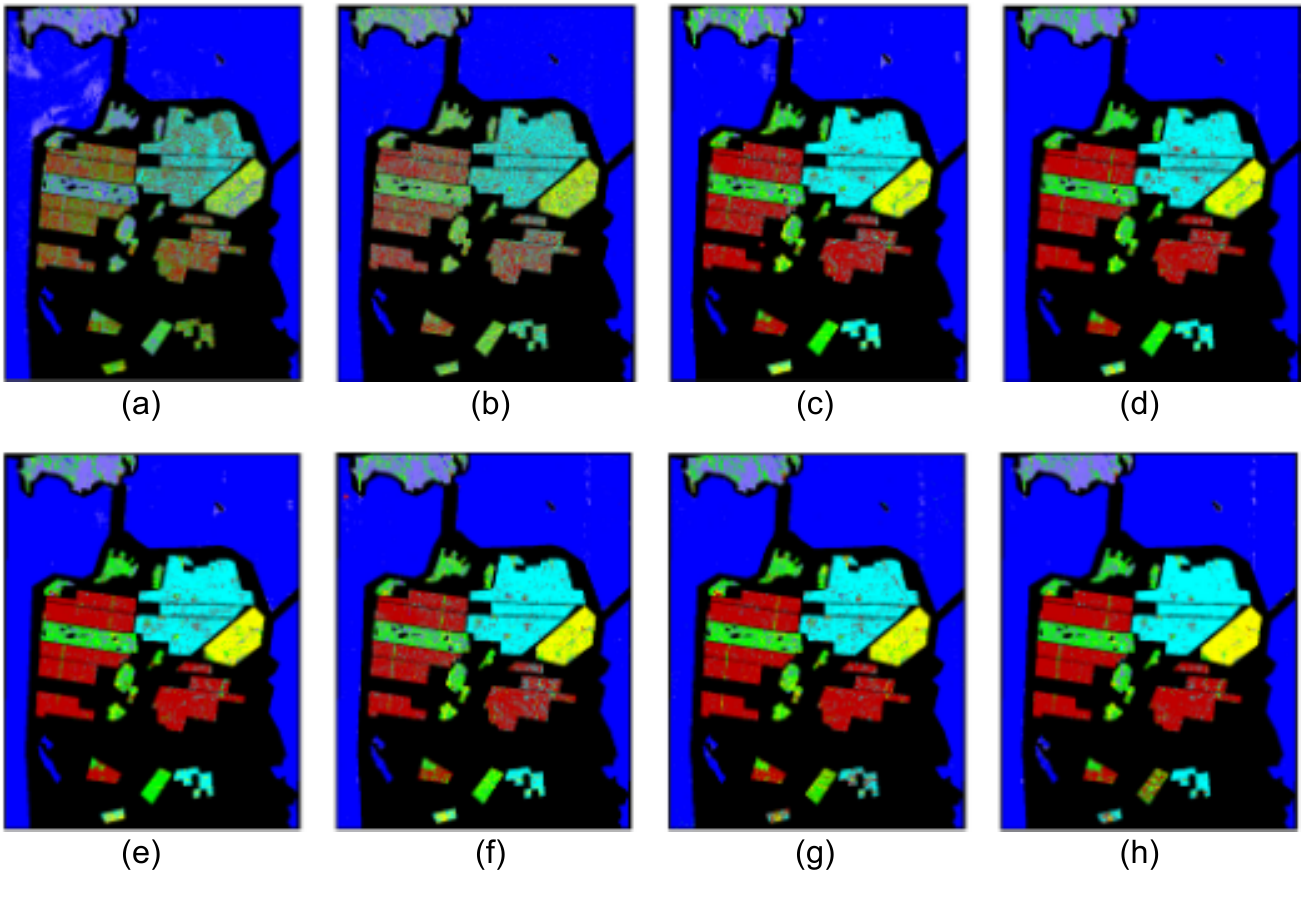}
	\caption{A visual  comparison, in terms of supervised classification of the entire map on the ALOS2 San Francisco dataset. Here, (a), (b), (c), (d), (e), (f), (g), and (h) shows the results obtained from Wishart, RBF-SVM, CV-CNN, 3D-CNN, PSENet, SF-CNN and ViT-PolSAR, respectively. Figure is from \cite{9658539}.}
	\label{fig:SAR_CLs_Qualit_comp}
\end{figure*}

\textbf{Classification:} Accurately classifying the target categories within SAR images is a challenging problem with numerous real-world applications. Recently, transformers have been explored for automatic interpretation and target recognition in SAR imagery. 
The work of \cite{9658539} explores vision transformers for polarimetric SAR (PolSAR) image classification. In this framework, the pixel values of the image patches are considered as tokens and the self-attention mechanism is employed to capture long-range dependencies followed by multi-layer perceptron (MLP) and learnable class tokens to integrate features. A contrastive learning technique is utilized within the framework to reduce the redundancies and perform the classification task. Fig.\ref{fig:SAR_PoISAR_Method} shows the overview of the framework and a qualitative comparison in terms of supervised classification is presented in Fig.\ref{fig:SAR_CLs_Qualit_comp}. s

Other than the aforementioned pure transformers-based approach, hybrid methods utilizing both CNNs and transformers also exist in literature. The work of \cite{9713848} introduces a global–local network structure (GLNS) framework that combines the merits of CNNs and transformers for SAR image classification. The proposed GLNS employs a lightweight
CNN along with an efficient vision transformer to capture both local and global features which are later fused to perform the classification task. Other than standard fully-supervised learning, transformers are also explored in the limited supervision regime such as, few-shot SAR image classification. Cai \textit{et al.} \cite{cai2022st} introduce a few-shot SAR classification approach, named ST-PN, where a spatial transformer network is utilized for performing spatial alignment on CNN-based features. 

\textbf{Segmentation and Detection:} Detection and segmentation in SAR imagery is vital for different applications such as, crop identification, target detection, and terrain mapping. In SAR imagery, segmentation can be challenging due to the appearance of speckles  which is a type of multiplicative noise that increases with the back-scattering radar magnitude. Among recent transformers-based approaches, the work of \cite{ke2022gcbanet} introduces a framework, named GCBANet, for SAR ship instance segmentation. Within the GCBANet framework, a global contextual block is employed to encode spatial holistic long-range dependencies. Furthermore, a boundary-aware box prediction technique is introduced to predict the boundaries of the ship. Xia \textit{et al.} \cite{xia2022crtranssar} introduce an approach, named CRTransSar, that combines the benefits of CNNs and transformers to capture both local and global information for SAR object detection. The proposed CRTransSar works by constructing a backbone with attention and convolutional blocks. A geospatial transformer framework is introduced in \cite{chen2022geospatial}, comprising the steps of image decomposition, multi-scale geo-spatial contextual attention and recomposition for detecting aircrafts in SAR imagery. A feature relation enhancement framework is proposed in \cite{zhang2022sfre} for aircraft detection in SAR imagery. The proposed framework adopts a fusion pyramid structure to combine features of different levels and scales. Further, a context attention enhancement technique is employed to improve the positioning accuracy in complex backgrounds. 

Other than ship and aircraft detection, the recent work of \cite{ma2021end} introduces a transformers-based framework for 3D detection of oil tank targets in SAR imagery. In this framework, the incidence angle is input to the transformer as a prior token followed by a feature description operator that utilizes scattering centers for refining the predictions. \\

\begin{table*}[h]
	\centering
	\caption{Overview of transformers-based approaches in SAR imaging. Here, we highlight methods for different SAR remote sensing tasks.}
	\resizebox{\textwidth}{!}{
	\begin{tabular}{V{3}l|c|c|c|P{7cm}V{3}} \hlineB{3}
		\rowcolor{mygray} \multicolumn{5}{|c|}{\textbf{Transformers in SAR Satellite Imagery}} \\ \hlineB{2}
		\rowcolor{mygray}	\textbf{Method}& \textbf{Task}& \textbf{Datasets}& \textbf{Metrics}& \textbf{Highlights} \\\hlineB{2}
			

			 ViT-PolSAR \cite{9658539} &Classification&\makecell{AIRSAR Flevoland \cite{norikane1992application},\\ ESAR Oberpfaffenhofen\cite{ESAR},\\ AIRSAR San Francisco\cite{sanfran},\\ ALOS2 San Francisco\cite{alos2}}&\makecell{AA,\\ OA,\\ Kappa}& Explores transformers, where self-attention is used to capture long-range dependencies followed by MLP for polarimetric SAR image classification.\\\hline
			 
			 GLNS \cite{9713848} &Classification&\makecell{Gaofen-3 SAR\cite{Gaofen3},\\ F-SAR\cite{FSAR}}&\makecell{AA,\\ OA,\\ Kappa}&Introduces a global–local network structure to exploit the merits of CNNs and transformers with local and global features that are fused to perform classification. \\\hline
			 
			 ST-PN \cite{cai2022st}&Classification&\makecell{MSTAR\cite{MSTAR}}&\makecell{Accuracy}&Proposes a spatial transformer network for spatial alignment of features extracted from CNNs for few-shot SAR classification. 

\\\hline

			 GCBANet \cite{ke2022gcbanet}&Segmentation &\makecell{SSDD\cite{8124934},\\ HRSID\cite{9127939}}&\makecell{AP}&Introduces a transformers-based approach with a global contextual block for capturing spatial holistic long-range dependencies and a boundary-aware prediction scheme for estimating the boundaries of ship. \\\hline
			 
			 CRTransSar \cite{xia2022crtranssar}&Detection &\makecell{SMCDD\cite{xia2022crtranssar},\\ SSDD\cite{8124934}}
			 &\makecell{Accuracy,\\ Recall,\\ mAP,\\ F1}& Proposes a backbone based on convolutional and attention blocks for capturing both local and global features.\\\hline
            
			 Geospatial Transformers \cite{chen2022geospatial}&Detection &\makecell{Gaofen-3\cite{Gaofen3}}
			 &\makecell{DR,\\ FAR}& 
Introduces a framework with multi-scale geo-spatial attention for aircraft detection in SAR imaging. \\\hline
			 
			SFRE-Net \cite{zhang2022sfre}&Detection &\makecell{Gaofen-3\cite{Gaofen3}}
			&\makecell{Precision,\\ Recall,\\ F1}& Introduces a feature relation enhancement architecture consisting of a fusion pyramid structure and a context attention enhancement technique. \\\hline

            3DET-ViT \cite{ma2021end}&Detection &\makecell{L1B SAR\cite{L1B}}
			&\makecell{AP,\\ AR,\\ Mean Offset}&
Proposes a transformers-based framework that takes incidence angle as a prior token with a feature description operator employing scattering centers for prediction refinement. 
\\\hline

			ID-ViT \cite{Perera22neural}&Despeckling&\makecell{Berkeley Segmentation Dataset\cite{937655}}
			&\makecell{PSNR,\\ SSIM}& Proposes a framework comprising an encoder to learn global dependencies among SAR image regions, where the network is trained using synthetic speckled data. 
\\\hline

			 CLT \cite{Huihui22neural}&Change Detection &\makecell{Brazil and Namibia datasets\cite{BrazilNamibia},\\ simulation data\cite{Huihui22neural}}&\makecell{KC}& 
Introduces a self-supervised contrastive representation learning method with a convolution-enhanced transformer to generate hierarchical representations for distinguishing changes from HR SAR images. \\\hline
			 
			 CF-ViT \cite{Fan22RS}&Image Registration&\makecell{MegaDepth\cite{MegaDepthLi18}}&\makecell{KC}& A CNN-transformers framework that first performs coarse registration on the down-sampled image, followed by registration of image pairs via a CNN-transformer module with the resulting point pair subsets integrated to obtain final global registration. \\\hline
			 
 		\end{tabular}
 		}
 	\label{tab:SAR}
\end{table*}
\subsection{Others}\label{MHSA}
Apart from SAR image classification, detection and segmentation, few works exist exploring transformers for other SAR imaging problems such as, image despeckling. 

\textbf{SAR Image Despeckling:} The aforementioned interpretation of SAR imaging is made challenging due to the degradation of images caused by a multiplicative noise known as speckle. Recently, transformers have been explored for SAR image despeckling. The work of \cite{Perera22neural} introduces a transformers-based framework comprising an encoder that learns global dependencies among various SAR image regions. The transformers-based network is trained in an end-to-end fashion with synthetic speckled data by utilizing a composite loss function. 

\textbf{Change Detection in SAR Images:} SAR images can be affected by imaging noise which presents challenges when detecting changes in high-resolution (HR) SAR data. Recently, a self-supervised contrastive representation learning technique has been proposed by \cite{Huihui22neural}, where hierarchical representations are constructed using a convolution-enhanced transformer to distinguish the changes from HR SAR images. A convolution-based module is introduced to enable interactions across windows when performing self-attention computations within local windows. 

\textbf{SAR Image Registration:} Several applications such as, change detection involves joint analysis and processing of multiple SAR images that are likely acquired in  different imaging conditions.  Thus, accurate SAR image registration is desired where the reference and the sensed images are registered. The recent work of \cite{Fan22RS} explores transformers for large-size SAR dense-matching registration. Here, a hybrid CNN-transformer is employed to register images under weak texture condition. First, coarse registration is performed via the down-sampled original SAR image. Then,  cluster centers of registration points are selected from the previous coarse registration step. Afterwards, the registration of image pairs are performed using a CNN-transformer module. Lastly, the resulting point pair subsets are integrated to achieve the final global transformation through RANSAC. 

In summary, we present a holistic overview of the existing transformers techniques in SAR imagery in Tab.\ref{tab:SAR}.

\section{Discussion and Conclusion}\label{VHR_discussion}
In this work, we presented a broad overview of transformers in remote sensing imaging: very-high resolution (VHR), hyperspectral and synthetic aperture radar (SAR). Within these different remote sensory imagery, we further discuss transformers-based approaches on a variety of tasks, such as classification, detection and segmentation. Our survey covers more than 60 transformers-based remote sensing research works in literature. We observed transformers to obtain favorable performance on different remote sensing tasks likely due to their capabilities to capture long-range dependencies along with their representation flexibility. Further, the public availability of several standard transformers architectures and backbones make it easier to explore their applicability in remote sensing imaging problems. 

\textbf{Open Research Directions:} As discussed earlier, most existing transformer-based recognition approaches employ backbones pre-trained on the ImageNet dataset. One exception is the work of \cite{Wang2022empirical} that explore pre-training vision transformers on a large-scale remote sensing dataset. However, in both cases the pre-training is performed in a supervised fashion. An open direction is to explore large-scale pre-training in a self-supervised fashion by taking into account an abundant amount of unlabeled remote sensing imaging data.

Our survey also shows that most existing approaches typically utilize a hybrid architecture where the aim is to combine the merits of convolutions and self-attention. However, transformers are typically known to have a higher computational cost to compute global self-attention. Several recent works have explored different improvements in the transformers design such as, reduced computational overhead \cite{CSWin}, efficient hybrid CNN-transformers backbones \cite{MobileViT} and unified architectures for image and video classification \cite{MViTv2}. Moreover, due to the utilization of more training data by transformers, there is a need to construct larger-scale datasets in remote sensing imaging. For most problems discussed in this work and especially in case of object detection, heavy backbones are typically utilized to achieve better detection accuracy. However, this significantly slows down the speed of the aerial detector. An interesting open direction is to design light-weight transformers-based backbones to classify detect oriented targets in remote sensing imagery.  Another open research direction is to explore the  adaptability of the transformers-based models to heterogeneous source of images such as, SAR and UAV (e.g., change detection).

In this survey, we also observe several existing approaches to utilize transformers in a plug-and-play fashion for remote sensing. This leads to the need of designing effective domain-specific architectural components and loss formulations to further boost the performance. Moreover, it is intriguing to study the adversarial feature space of vision transformers models that are pre-trained on remote sensing benchmarks and their transferability.

\bibliographystyle{ieeetr}
\bibliography{bibliography}

\vfill

\end{document}